\documentclass[11pt]{article}
\usepackage[preprint]{acl}
\usepackage[title]{appendix}
\usepackage{booktabs}
\usepackage{tabularx}
\usepackage{array}
\usepackage{chngcntr}
\usepackage{graphicx}
\usepackage{enumitem}
\usepackage{amsmath}
\usepackage{multirow}
\usepackage{algorithm}
\usepackage{algpseudocode}
\usepackage{listings}
\usepackage{tikz}
\usetikzlibrary{
  arrows.meta,
  positioning,
  calc,
  shapes.geometric
}

\tikzset{
  causalnode/.style={
    draw,
    ellipse,
    minimum height=7mm,
    minimum width=18mm,
    inner xsep=7pt,
    align=center,
    font=\small
  },
  causal/.style={
    -{Stealth[length=2.2mm,width=1.5mm]},
    semithick
  }
}
\counterwithin{table}{section}
\counterwithin{figure}{section}

\newcounter{prompt}[section]

\usepackage{microtype}

\usepackage{iftex}
\usepackage{fontspec}
\usepackage[english,bidi=default]{babel}

\babelfont{rm}[
  BoldFont = TeXGyreTermesX-Bold.otf,
  ItalicFont = TeXGyreTermesX-Italic.otf,
  BoldItalicFont = TeXGyreTermesX-BoldItalic.otf
]{TeXGyreTermesX-Regular.otf}

\ifXeTeX
  \usepackage{xeCJK}
  \usepackage[kanakinsoku]{zxjatype}

  \setCJKsansfont[
    BoldFont = HaranoAjiGothic-Bold.otf
  ]{HaranoAjiGothic-Regular.otf}

  \setCJKmonofont{HaranoAjiGothic-Regular.otf}

\else
  \ifLuaTeX
    \usepackage{luatexja-fontspec}

    \setmainjfont[
      BoldFont = HaranoAjiMincho-Bold.otf
    ]{HaranoAjiMincho-Regular.otf}

    \setsansjfont[
      BoldFont = HaranoAjiGothic-Bold.otf
    ]{HaranoAjiGothic-Regular.otf}

    \setmonojfont{HaranoAjiGothic-Regular.otf}

  \else
    \PackageError{}{Use XeLaTeX or LuaLaTeX}{This document requires XeLaTeX or LuaLaTeX for Japanese text.}
  \fi
\fi

\title{MaSCoD: A Multi-Agent Framework for
Structural-Context-Guided Candidate Causal Graph Generation}

\author{Yudai Nakada\thanks{These authors contributed equally to this work.} \\
SCSK Corporation \\
Tokyo, Japan \\
\texttt{yu.nakada@scsk.jp}\And
  Yuichiro Nishiura\footnotemark[1] \\
  SCSK Corporation \\
  Tokyo, Japan \\
\texttt{y.nishiura@scsk.jp}\And
  Jin Michael Splichal \\
  SCSK Corporation \\
  Tokyo, Japan \\
\texttt{J.Splichal@scsk.jp}
}

\begin{document}
\maketitle
\begin{abstract}
Large language models (LLMs) have been applied to causal discovery,
but candidate-graph generation rarely treats premature omission of
potentially relevant causal relations as an explicit design objective.
We propose MaSCoD, a multi-agent framework that organizes candidate
third variables and local structural patterns before direct-edge
judgment.
We evaluate MaSCoD on Auto-MPG, DWD, and Sachs using GPT-5.4 as the
primary backbone and GPT-4o for replication.
MaSCoD exhibits a dataset- and backbone-dependent
retention--selectivity profile rather than uniform superiority.
Across all six dataset--backbone settings, Full, which supplies structural hypotheses before direct-edge judgment, achieved
higher mean Recall and F1 than No Phase~1, which instead constructs them within the judgment procedure, while also
increasing false-positive rates.
Additional reference-edge retention over all evaluated
baselines was observed on DWD with GPT-5.4 and on
Sachs with GPT-4o, rather than uniformly across settings.
Partial ablations showed that supplying both information
components did not always outperform supplying only one.
For GPT-5.4, stage-wise analysis showed that the
Full--No Phase~1 retention gap was already present after
direct-edge judgment, while reconciliation introduced
additional reference-edge loss for Full on Sachs.
These findings support structural pre-organization as
an explicit design and evaluation target for omission
control and motivate evaluating context construction
jointly with its utilization in judgment.
\end{abstract}
\section{Introduction}
\label{sec:introduction}
Causal discovery is central to empirical research in epidemiology, economics, and the social sciences, particularly when randomized experiments are infeasible, costly, or ethically constrained. Statistical approaches such as PC, GES, LiNGAM, and NOTEARS infer candidate structures from observational data under specific assumptions \cite{Glymour2019,chickering2002optimal,shimizu2006,Zheng2018}. These methods do not necessarily recover the true causal graph and may be affected by latent confounding, measurement error, or selection bias. This matters because causal discovery often supports downstream tasks such as validating causal pathways, selecting adjustment variables, and identifying intervention targets \cite{Zanga2022,Takayama2025}.At the candidate-generation stage, false positives and false negatives
can have asymmetric consequences: an included edge may be removed
downstream, whereas an omitted causal relation may never be examined.
We therefore treat retention of potentially relevant causal relations
as an explicit design requirement alongside overall graph quality.

Candidate-graph formation also relies on domain knowledge concerning temporal ordering, theoretical regularities, and background mechanisms that may be weakly reflected in observed associations. Such knowledge can be incorporated as prior constraints or used to assess estimated structures, but collecting and applying it systematically is labor-intensive \cite{Hasan2022,Takayama2025}. Recent work has consequently explored large language models (LLMs) as scalable tools for knowledge elicitation and causal judgment. LLMs have been used to provide prior knowledge for statistical discovery, support graph estimation, and directly generate causal graphs \cite{Jiralerspong2024,Takayama2025,Le2025}. Most existing approaches, however, primarily target overall graph quality, efficiency, or judgment reliability rather than premature omission during candidate-graph generation.

Prior findings indicate that LLM-based causal judgments depend on how structural context is presented. Triplet prompts can yield more stable causal ordering than pairwise prompts, while decomposed procedures have been proposed to distinguish direct from indirect causation by identifying mediators first \cite{Vashishtha2025,Cai2025}. Motivated by these findings, we distinguish the construction
of structural hypotheses from their utilization in direct-edge
judgment.
Our input-design hypothesis is that explicitly organizing
third-variable identities and local structural patterns before
judgment can reduce reference-edge omissions relative to
requiring these hypotheses to be constructed within the same
judgment procedure.
This motivates evaluating structural pre-organization as
a design choice for omission control, without assuming
an improvement in the LLM's underlying reasoning ability.

We propose MaSCoD, an omission-oriented multi-agent framework for structural-context-guided candidate causal graph generation. For each ordered variable pair, Phase 1 elicits and selects candidate third variables and assigns them representative Fork, Chain, or Collider patterns. Phase 2 evaluates whether a direct edge should be retained under these auxiliary structural hypotheses. Phase 3 integrates the ordered-pair judgments, resolves bidirectional edges, and attempts BIC-based post-hoc cycle reconciliation using observational data. The resulting graph is treated as a candidate structure for downstream examination rather than as a definitive estimate of the true causal graph. MaSCoD's novelty lies not in using third variables per se, but in organizing them before direct-causality judgment under the explicit objective of omission control.

We address two research questions: (RQ1) to what extent MaSCoD reduces reference-edge omissions relative to statistical and LLM-based baselines, and what false-positive burden accompanies its retention performance across datasets and backbone settings; and (RQ2) how configurations that differ in the structural hypotheses supplied before direct-edge judgment compare in reference-edge retention and false-positive inclusion, and where in the pipeline reference-edge omissions arise. Our contributions are threefold.
First, we frame structural pre-organization as a design
problem for omission-oriented candidate causal graph
generation, distinguishing which auxiliary hypotheses
are made explicit before direct-edge judgment from how
they are subsequently used.
Second, we instantiate this distinction in MaSCoD through
third-variable and pattern organization, multi-persona
judgment, and graph-level reconciliation.
Third, we empirically characterize the implemented design
conditions through baseline comparisons, information
ablations, and stage-wise retention analysis, examining
both the retention benefits of pre-organization and
the limits of those benefits.
\section{Related Work}

LLMs have been used to provide background knowledge and
conditional-independence judgments for statistical causal discovery
\cite{Long2023,Cohrs2023,Takayama2025}, to generate graph hypotheses
directly \cite{Jiralerspong2024,Roy2025}, and in hybrid procedures that
combine LLM reasoning with downstream estimation or refinement
\cite{Le2025,Shen2025}.
ASoT \cite{Meier2025} further uses hierarchical decomposition and
multi-agent consensus for agentic causal discovery.
Across these approaches, recurring design strategies include enriching
structural context, decomposing difficult judgments, aggregating
multiple responses, and reconciling local decisions at the graph
level.
Prior work primarily emphasizes graph quality, efficiency, or
reliability; premature omission during candidate-graph formation is
rarely treated as an explicit design objective.

Two findings particularly motivate our design.
Vashishtha et al.\ \cite{Vashishtha2025} report that triplet-based
querying can improve causal ordering relative to pairwise prompting,
while Cai et al.\ \cite{Cai2025} separate mediator identification from
direct- and indirect-causality judgment.
These results suggest that structural context and staged reasoning can change which relations remain plausible during judgment. However, merely presenting a candidate third variable does not ensure that the corresponding structural alternative is explicitly considered during direct-edge judgment. By assigning each candidate a local structural role and treating it as a thought-experiment condition, MaSCoD encourages the model to evaluate the direct-edge hypothesis relative to a specified structural context rather than as an isolated pairwise question. MaSCoD builds on this design principle by organizing candidate third variables and local structural patterns before direct-edge judgment, followed by multi-persona aggregation and graph-level reconciliation.
\section{Method}
\label{sec:method}

\subsection{Overview}
\label{subsec:method-overview}

Let $V=\{X_1,\ldots,X_n\}$ denote the variable set, let $d_i$ denote the natural-language description of $X_i$, and let $D$ denote the observational data. MaSCoD produces an adopted directed graph $G^*$ through three stages: structural-context construction, direct-edge judgment, and structural reconciliation. Here, structural context refers to auxiliary third-variable
identities and their local structural patterns, rather than
to the entire prompt or to independently verified causal facts.
Phase~1 constructs and selects these hypotheses, while
Phase~2 utilizes them through candidate-level judgment and,
when needed, cross-candidate selection.
This construction--utilization distinction concerns local
edge judgment; Phase~3 separately reconciles the resulting
graph. Phases~1 and~2 use variable names and descriptions to process all $n(n-1)$ ordered pairs; the observational data are used only in the
BIC-based reconciliation step in Phase~3. We evaluate $(X,Y)$ and $(Y,X)$ separately because relevant third variables and structural hypotheses are direction-specific, and we resolve conflicts only after the two directions have been judged independently.

Each LLM-based stage uses three personas with different domain specializations. Within each stage, the personas receive the same task definition and pair-specific information but contribute different interpretive perspectives; majority rules are used to consolidate their judgments. Because an ordered pair may be evaluated under multiple third-variable candidates, the effective cost exceeds the number of ordered pairs. The framework is
therefore intended primarily for small-to-medium variable sets. Persona definitions, prompts, and detailed pseudocode are provided in Appendices~C--I.
  
\begin{figure*}[t]
\centering
\includegraphics[width=\textwidth]{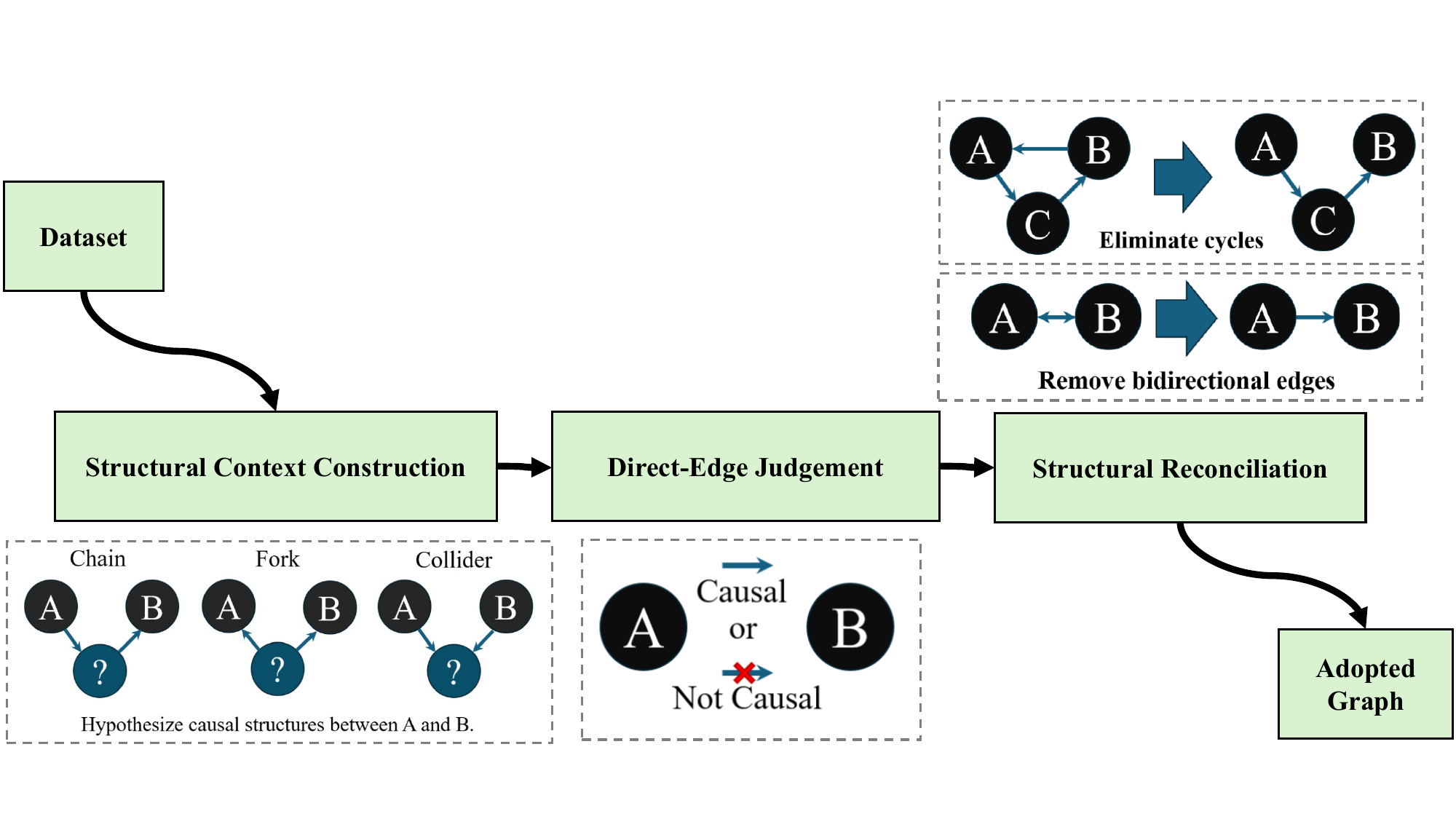}
\caption{Overview of the proposed candidate-graph-generation framework.
The method takes variable names, natural-language variable descriptions,
and observational data as input, and produces an adopted candidate graph
through three stages. Phase~1 elicits and selects candidate third
variables and structural patterns for each ordered variable pair.
Phase~2 estimates whether a direct causal relation should be retained
under the third-variable context generated in Phase~1. Phase~3
integrates the resulting ordered-pair judgments, resolves bidirectional
edges, and applies post-hoc cycle removal. When cycle removal completes,
the adopted output is a DAG; otherwise, the graph obtained after
bidirectional-edge resolution is retained as the adopted estimate.}
\label{fig:mascod-overview}
\end{figure*}

\subsection{Phase 1: Structural-Context Construction}
\label{subsec:phase1}

For each ordered pair $(X,Y)$, Phase~1 constructs a candidate set $C_{XY}=\{(Z_j,r_j)\}$, where $Z_j$ is a third variable and $r_j$ is one of three representative local patterns: Fork $(X \leftarrow Z \rightarrow Y)$, Chain $(X \rightarrow Z \rightarrow Y)$, or Collider $(X \rightarrow Z \leftarrow Y)$. These patterns provide a compact scaffold for comparing structural alternatives and are not intended as an exhaustive causal taxonomy. The
candidates are auxiliary hypotheses for Phase~2 and are never inserted directly into the adopted graph.

Phase~1 follows an NGT-inspired generate-and-narrow procedure \cite{delbecq1975group,cantrill1996delphi}. First, each persona independently proposes
third variables and assigns a pattern to each candidate using the variable descriptions, temporal ordering, physical regularities, and domain
knowledge. Observed variables are preferred, but unobserved variables may be proposed when needed for explanation.The three candidate lists are then pooled, and duplicate or semantically similar candidates are merged using embeddings of their candidate-name
strings. When a deduplicated cluster contains different pattern
assignments, the most frequent pattern among its member records is
retained.

Each persona next scores every deduplicated candidate from 1 to 10 according to its importance for judging $X \rightarrow Y$. The scores are
aggregated and presented to the personas, which independently decide whether each candidate should be passed to Phase~2. A candidate is retained when the proportion of valid Yes votes is at
least 0.6; when all three persona votes are valid, this is equivalent
to requiring at least two Yes votes. This procedure separates candidate generation, comparison, and acceptance so that Phase~2 receives a structured set of alternatives rather than an unfiltered free-form output.

\subsection{Phase 2: Direct-Edge Judgment}
\label{subsec:phase2}

Phase~2 evaluates whether $X \rightarrow Y$ should be retained under the auxiliary hypotheses in $C_{XY}$. The supplied third-variable pattern is
treated as a thought-experiment condition rather than as an independently verified causal fact. For each candidate $(Z,r)$, the personas consider
temporal ordering, known mechanisms, and the expected change in $Y$ under a hypothetical change in $X$. Under Fork and Chain hypotheses, they assess
whether an effect of $X$ on $Y$ would remain when $Z$ is conceptually held fixed; under a Collider hypothesis, they assess the relation without conditioning on $Z$.

Candidate-level judgments use seven labels: \texttt{Fork}, \texttt{Fork+Direct}, \texttt{Chain}, \texttt{Chain+Direct}, \texttt{Collider},
\texttt{Collider+Direct}, and \texttt{Invalid}. Labels containing \texttt{+Direct} retain $X \rightarrow Y$, whereas the corresponding labels without
\texttt{+Direct} reject the direct edge under the supplied structural hypothesis. \texttt{Invalid} rejects the candidate-pattern hypothesis itself and does not indicate support for $Y \rightarrow X$. When a structural pattern is specified, the available
candidate-level labels are restricted to the two labels
associated with that pattern plus \texttt{Invalid};
when the pattern must be inferred, all seven labels
are available.

Judgment follows a Delphi-inspired procedure with at most two rounds \cite{Dalkey1963,linstone1975delphi}. In the first round, the
three personas respond independently. If at least two select the same label, that label is adopted. Otherwise, a structured summary of the first-round
judgments is returned to the personas for reassessment. If no majority emerges in the second round, the candidate-level result is mapped to the absence of a direct edge.

When Phase~1 retains multiple candidates, the resulting labels and summarized reasons are compared, and the personas select the most plausible and
explainable candidate-level judgment by majority vote. If only one candidate is retained, its result becomes the pair-level output directly. If $C_{XY}$ is empty, a no-predefined-candidate prompt is used, requiring the model to hypothesize a relevant third variable and pattern within the same judgment procedure.

\subsection{Phase 3: Structural Reconciliation}
\label{subsec:phase3}

The final Phase~2 labels are first mapped into an intermediate directed graph: labels containing \texttt{+Direct} add the corresponding edge, while
labels without \texttt{+Direct} and \texttt{Invalid} do not.Because opposite directions are evaluated independently, the intermediate graph may contain both \(X\rightarrow Y\) and \(Y\rightarrow X\). For each such conflict, the three personas compare the two surviving directional hypotheses. Each persona receives the descriptions of \(X\) and \(Y\) and the stored reason supporting each surviving direction.
The Phase~1 third-variable candidates are not presented again at this
stage, and the Phase~2 final labels themselves are not included in the
direction-selection prompt. Each persona is required to choose exactly one of \(X\rightarrow Y\) and \(Y\rightarrow X\). The prompt is coupled with a strict structured-output schema that permits only these two values. The direction selected by at least two of the three personas is retained, and the opposite direction is removed. Thus, when all three responses are successfully parsed, the bidirectional conflict is always reduced to one directed edge. If directed cycles remain after bidirectional-edge resolution, MaSCoD applies a post-hoc BIC-based reconciliation procedure using the observational data. It enumerates acyclic candidate graphs obtained by removing edges sufficient to break the detected cycles, fits each candidate under a linear-Gaussian structural-equation scoring assumption, and adopts the candidate with the lowest BIC. If this procedure produces no cycle-resolution output, the graph obtained after bidirectional-edge resolution is retained and may still contain directed cycles.
\section{Experimental Setup}
\label{sec:experimental-setup}

\paragraph{Datasets and reference graphs.}
We evaluate MaSCoD on Auto-MPG, DWD, and Sachs
\cite{quinlan1993autompg,mooij2016distinguishing,sachs2005causal},
following the dataset configuration and reference graphs used by
\citeauthor{Takayama2025}
Table~\ref{tab:datasets} summarizes the three settings.
The adopted reference graphs contain 5, 6, and 19 directed edges for
Auto-MPG, DWD, and Sachs, respectively.
All variables are continuous and are standardized before applying the
statistical baselines; the same standardized observations are used for
the BIC-based reconciliation in Phase~3.
For the LLM-based procedures, we provide only variable names and
author-written descriptions that explain variable meaning without
explicitly stating target edge directions.
The complete variable descriptions and adopted reference graphs are
provided in the appendix.

\begin{table}[t]
    \centering
    \small
    \setlength{\tabcolsep}{3.5pt}
    \begin{tabular}{lrrrr}
        \toprule
        Dataset & Vars. & Obs. & Pairs & Ref. edges \\
        \midrule
        Auto-MPG & 5  & 392   & 20  & 5  \\
        DWD      & 6  & 349   & 30  & 6  \\
        Sachs    & 11 & 7,466 & 110 & 19 \\
        \bottomrule
    \end{tabular}
    \caption{Datasets used in the main experiments. Pairs denotes
    ordered non-self variable pairs, and Ref. edges denotes the number
    of directed edges in the adopted reference graph.}
    \label{tab:datasets}
\end{table}

\paragraph{Baselines.}
We compare MaSCoD with three statistical causal-discovery
methods---PC, Exact Search, and DirectLiNGAM---and two LLM-based
baselines, LLM-KBCI and MAC
\cite{book,Silander2006,shimizu2011directlingam,Takayama2025,Le2025}.
The statistical methods are evaluated independently of the LLM
backbone. For the LLM-based comparisons, MaSCoD, LLM-KBCI, and MAC
use the same backbone within each comparison setting.

\paragraph{Metrics and repeated runs.}
The main text reports F1, Recall, and false-positive rate (FPR).
Recall directly evaluates the omission-control objective.
We interpret it together with FPR to characterize the
inclusion of non-reference edges, and report F1 as
a summary of Precision and Recall.
All metrics are defined relative to the adopted
reference graphs.
Full results for normalized Hamming distance (NHD), Precision, and
false-negative rate (FNR) are reported in the appendix.
For an $n$-variable graph, NHD is computed as
\[
    \mathrm{NHD} = \frac{FP+FN}{n^2}.
\]
For repeated experiments, each metric is computed separately for each
run and then summarized by the mean and sample standard deviation.

We use GPT-5.4 (\texttt{gpt5.4-2026-03-05}) as the primary LLM
backbone and GPT-4o (\texttt{gpt-4o-2024-11-20}) as a second backbone
for replication.
Both models were accessed through Azure OpenAI Service.
We repeat the method comparisons and Phase~1 ablations
with GPT-4o to examine whether the observed tendencies
persist under a second backbone configuration.
This comparison evaluates the pipeline under different
backbone settings rather than changing only the
judgment model while holding the constructed context fixed.
Each LLM-based method and ablation condition is run independently
three times per dataset.
The statistical baselines are evaluated once under the adopted
configuration because they do not involve stochastic LLM generation.

For GPT-5.4, the temperature was set to 1.0 and the maximum output
length to 1,024 tokens; other generation parameters were left at the
service defaults.

For GPT-4o, the temperature was set to 0.2 and the maximum output
length to 512 tokens; top-$p$ was left at the service default and
random seeds were not fixed.

We did not perform hyperparameter search or tune generation parameters
on the evaluation datasets; the settings above were fixed for each
backbone throughout the corresponding experiments.

\paragraph{Phase~1 ablations.}
To assess the role of structural pre-organization, we compare four
information conditions.
In \textsc{Full}, Phase~2 receives both the identity of the selected
third variable and its structural pattern.
In \textsc{Identity only}, the third-variable identity is supplied but
its structural pattern is withheld, requiring Phase~2 to infer the
relevant Fork, Chain, or Collider relation.
In \textsc{Pattern only}, the structural pattern is supplied while the
concrete identity of the third variable is withheld, requiring the
model to hypothesize a plausible variable consistent with that pattern.
In \textsc{No Phase 1}, neither the third-variable identity nor the
structural pattern is supplied; Phase~2 must hypothesize both before
judging whether the direct edge should be retained.

For each dataset--backbone setting, Full was evaluated
in three runs, each generating its own Phase~1 outputs
and executing Phase~2 using those outputs.
The partial-information conditions instead reused
the Phase~1 outputs from Full run~1.
This source run was chosen because it was the first
in execution order, not on the basis of its generated
content or evaluation performance.

Identity only was constructed by masking the
structural-pattern information in these outputs,
whereas Pattern only was constructed by masking
the third-variable identity.
For each partial-information condition, Phase~2
was executed three times using the same fixed
masked inputs.
The Phase~1 outputs from Full runs~2 and~3 were
not used in the partial-information experiments.

Accordingly, Identity only and Pattern only share
one source Phase~1 realization per dataset--backbone
setting.
Their means and sample standard deviations summarize
downstream repetitions conditional on that realization.
In contrast, the Full results aggregate three runs
with separately generated Phase~1 outputs.
Comparisons of the partial conditions with the
three-run Full mean are therefore not matched across
Phase~1 realizations.
The downstream prompts and judgment requirements
therefore differ across conditions.
Accordingly, the ablation compares implemented
construction--utilization configurations rather than
isolating the effect of context content supplied to
an otherwise identical judgment procedure.

\paragraph{Stage-wise retention analysis.}
For the primary GPT-5.4 experiments, we additionally compare
\textsc{Full} and \textsc{No Phase 1} by tracing every reference edge
through the direct-edge judgment and structural-reconciliation stages.
A reference edge is counted as retained after Phase~2 when its
pair-level label contains \textsc{+Direct}.
It is counted as retained in the final graph only when the same
directed edge remains in the adopted output.
An edge that is retained after Phase~2 but absent from the adopted
graph is counted as a loss during Phase~3 reconciliation.
These proportions are computed separately for each of the three runs
and then summarized across runs.
\section{Results}
\label{sec:results}

We report results for the two research questions introduced in
Section~\ref{sec:introduction}. GPT-5.4 is treated as the
primary backbone, and GPT-4o is used to examine whether the observed
tendencies persist under a second LLM backbone. Unless otherwise
noted, results for LLM-based methods are averages over three runs.
Because Recall and FPR directly expose the trade-off between retaining
reference edges and introducing spurious edges, the main-text tables
report these metrics together with F1. Complete results including
Precision, FNR, NHD, and sample standard deviations are provided in
Appendix~\ref{app:full-results}.

\subsection{RQ1: Omission Control Relative to Existing Methods}
\label{sec:results_rq1}

\begin{table*}[t]
\centering
\small
\setlength{\tabcolsep}{3.1pt}
\begin{tabular}{llccc@{\hspace{7pt}}ccc@{\hspace{7pt}}ccc}
\toprule
& &
\multicolumn{3}{c}{Auto-MPG} &
\multicolumn{3}{c}{DWD} &
\multicolumn{3}{c}{Sachs} \\
\cmidrule(lr){3-5}\cmidrule(lr){6-8}\cmidrule(lr){9-11}
Backbone & Method &
F1$\uparrow$ & R$\uparrow$ & FPR$\downarrow$ &
F1$\uparrow$ & R$\uparrow$ & FPR$\downarrow$ &
F1$\uparrow$ & R$\uparrow$ & FPR$\downarrow$ \\
\midrule
\multirow{3}{*}{--}
& PC             & .167 & .200 & .400 & .143 & .167 & .292 & .444 & .526 & .176 \\
& DirectLiNGAM   & .143 & .200 & .533 & .211 & .333 & .458 & .364 & .526 & .286 \\
& Exact Search   & .615 & .800 & .267 & .353 & .500 & .333 & .385 & .526 & .253 \\
\midrule
\multirow{3}{*}{GPT-5.4}
& MaSCoD         & .603 & .800 & .289 & .708 & .944 & .181 & .417 & .526 & .209 \\
& LLM-KBCI       & .615 & .800 & .267 & .488 & .667 & .264 & .582 & .526 & .059 \\
& MAC            & .687 & .800 & .178 & .586 & .556 & .083 & .443 & .316 & .022 \\
\midrule
\multirow{3}{*}{GPT-4o}
& MaSCoD         & .571 & .800 & .333 & .586 & .667 & .153 & .456 & .684 & .275 \\
& LLM-KBCI       & .684 & .867 & .222 & .601 & .667 & .139 & .395 & .298 & .044 \\
& MAC            & .605 & .800 & .289 & .543 & .556 & .125 & .390 & .281 & .033 \\
\bottomrule
\end{tabular}
\caption{Comparison with statistical and LLM-based causal-discovery
methods. R denotes Recall. Statistical methods are backbone-independent.
LLM-based entries are three-run means. Complete metrics and dispersion
estimates are reported in Appendix~\ref{app:full-results}.}
\label{tab:main_results}
\end{table*}

Table~\ref{tab:main_results} compares MaSCoD with the
backbone-independent statistical baselines and the LLM-based
baselines evaluated with the same backbone.
MaSCoD achieved the highest reported Recall, including ties,
in five of the six dataset--backbone settings.
It exceeded every baseline in Recall on DWD with GPT-5.4
(0.944) and on Sachs with GPT-4o (0.684), also achieving
the highest F1 in these two settings
(0.708 and 0.456, respectively).

The accompanying differences in FPR depended on the comparison.
On DWD with GPT-5.4, MaSCoD achieved higher Recall and lower
FPR than LLM-KBCI, but higher Recall and higher FPR than MAC.
On Sachs with GPT-4o, its Recall advantage over the two
LLM-based baselines was accompanied by higher FPR
(0.275, compared with 0.044 for LLM-KBCI and 0.033 for MAC).
In the remaining four settings, at least one baseline achieved
equal or higher Recall with lower FPR.
Thus, a retention advantage over all baselines was observed
in two settings, rather than uniformly across the evaluated
settings, and its false-positive implications depended on
the comparison method.

\subsection{RQ2: Phase 1 and Its Information Components}
\label{sec:results_rq2}

We compare four Phase~1 information conditions: Full, Pattern only,
Identity only, and No Phase~1. The partial-information conditions reuse the Phase~1 outputs
from a single Full run---the first of the three---in each
dataset--backbone setting.
Across both backbones and all three datasets, Full yields higher Recall
than No Phase~1, but also higher FPR.
Thus, Phase~1 shifts the retention--selectivity operating point rather
than uniformly improving graph quality.

The partial ablations show no backbone-independent ordering between
third-variable identity and structural-pattern information.
Under GPT-5.4, Identity only yields higher Recall than Pattern only on
all three datasets, whereas this ordering disappears or reverses under
GPT-4o.
Moreover, Full does not uniformly maximize F1 or Recall. These comparisons with Full are descriptive rather than
matched-context comparisons, because the Full mean
aggregates three separately generated Phase~1 realizations,
whereas both partial conditions reuse only the first.
These results indicate that the contribution of each Phase~1
information component depends on the dataset, backbone, and their
combination in downstream judgment.
For GPT-5.4, the stage-wise analysis further shows that the
Full--No Phase~1 difference largely emerges by the direct-edge
judgment stage.
Under Full, reference-edge retention from Phase~2 to the final graph
was 0.800$\rightarrow$0.800 on Auto-MPG,
0.944$\rightarrow$0.944 on DWD, and
0.702$\rightarrow$0.526 on Sachs.
Under No Phase~1, the corresponding values were
0.400$\rightarrow$0.400, 0$\rightarrow$0, and
0.070$\rightarrow$0.070.
Thus, no additional reference-edge loss occurred during Phase~3 for
No Phase~1 in these experiments, whereas the Full condition on Sachs
lost additional edges during structural reconciliation.
Complete stage-wise results are reported in
Appendix~\ref{app:edge-diagnostics}.

\begin{table*}[t]
\centering
\small
\setlength{\tabcolsep}{3.1pt}
\begin{tabular}{llccc@{\hspace{7pt}}ccc@{\hspace{7pt}}ccc}
\toprule
& &
\multicolumn{3}{c}{Auto-MPG} &
\multicolumn{3}{c}{DWD} &
\multicolumn{3}{c}{Sachs} \\
\cmidrule(lr){3-5}\cmidrule(lr){6-8}\cmidrule(lr){9-11}
Backbone & Condition &
F1$\uparrow$ & R$\uparrow$ & FPR$\downarrow$ &
F1$\uparrow$ & R$\uparrow$ & FPR$\downarrow$ &
F1$\uparrow$ & R$\uparrow$ & FPR$\downarrow$ \\
\midrule
\multirow{4}{*}{GPT-5.4}
& Full          & .603 & .800 & .289 & .708 & .944 & .181 & .417 & .526 & .209 \\
& Pattern only  & .442 & .533 & .289 & .535 & .611 & .167 & .393 & .421 & .147 \\
& Identity only & .615 & .800 & .267 & .636 & .833 & .194 & .480 & .702 & .253 \\
& No Phase 1    & .433 & .400 & .156 & .000 & .000 & .083 & .117 & .070 & .026 \\
\midrule
\multirow{4}{*}{GPT-4o}
& Full          & .571 & .800 & .333 & .586 & .667 & .153 & .456 & .684 & .275 \\
& Pattern only  & .571 & .800 & .333 & .571 & .667 & .167 & .371 & .719 & .451 \\
& Identity only & .546 & .800 & .378 & .571 & .667 & .167 & .379 & .474 & .245 \\
& No Phase 1    & .530 & .600 & .222 & .444 & .333 & .042 & .426 & .579 & .238 \\
\bottomrule
\end{tabular}
\caption{Phase~1 ablation. Full supplies both third-variable identity
and structural-pattern information. All entries are three-run means.
For the GPT-4o Sachs partial-information conditions, cycle resolution
timed out in all Pattern-only runs and in one Identity-only run; the
post-direction-resolution Phase~3 graph was evaluated in those cases,
following the pipeline fallback rule.}
\label{tab:ablation_results}
\end{table*}

\section{Discussion}
\label{sec:discussion}

\paragraph{Omission control relative to existing methods.}
The baseline comparisons identify settings in which MaSCoD
provides additional reference-edge retention.
Its clearest retention advantages over all evaluated baselines
occurred on DWD with GPT-5.4 and on Sachs with GPT-4o.
MaSCoD also achieved the highest reported F1 in these settings,
showing that the retention gains were not accompanied by
lower F1.
The false-positive implications, however, depended on the
comparison method.
On DWD with GPT-5.4, MaSCoD improved both Recall and FPR
relative to LLM-KBCI, whereas its advantage over MAC involved
higher Recall and higher FPR.
On Sachs with GPT-4o, the retention advantage over the
LLM-based baselines came with higher FPR.
In the other four settings, a baseline achieved equal or higher
Recall with lower FPR.
Thus, matching the highest Recall does not by itself establish
a comparative advantage.
The results support additional retention in specific settings,
rather than a general preference for MaSCoD over existing methods.

\paragraph{Structural context construction and utilization.}
The ablations motivate a construction--utilization perspective
on omission-oriented candidate-graph generation.
Phase~1 makes selected third-variable hypotheses and their
structural interpretations explicitly available before
direct-edge judgment, whereas Phase~2 evaluates whether a
direct edge should be retained under those hypotheses.
No Phase~1 does not eliminate structural reasoning:
it instead requires the judgment procedure to construct
both the third-variable hypothesis and its structural pattern.
The design distinction is therefore between pre-organizing
structural alternatives and constructing them within the
same procedure that decides whether to retain an edge.

Within each backbone setting, Full achieved higher mean
Recall and F1 than No Phase~1 on all three datasets,
while also increasing FPR.
These results support treating structural pre-organization
as an explicit object of design and evaluation, rather than
attributing the observed candidate-graph performance to
the backbone alone.

Taken together, these findings motivate evaluating the
construction of structural context jointly with the procedure
that uses it, under the explicit objective of omission control.
They do not, however, identify an isolated interaction between
context content and the judgment model.
The information conditions also differ in their prompts and
judgment requirements, while changing the backbone changes
both context generation and subsequent judgment.
We therefore interpret the findings as evidence about
implemented construction--utilization configurations,
not as a mechanistic account of improved LLM reasoning.

\paragraph{Stage-wise constraints on omission control.}
The stage-wise analysis distinguishes two requirements:
retaining reference edges through local judgment and
preserving those edges during structural reconciliation.
In the implemented pipeline, Phase~3 only selects among
existing directions and removes edges.
If $E_{\mathrm{P2}}$ and $E_{\mathrm{final}}$ denote the
directed-edge sets after Phase~2 and in the adopted graph,
respectively, then
$E_{\mathrm{final}} \subseteq E_{\mathrm{P2}}$.
Consequently,
\[
R_{\mathrm{final}} \leq R_{\mathrm{P2}},
\]
where the retention measures are defined in
Appendix~\ref{app:edge-diagnostics}.
Thus, edges omitted before reconciliation cannot be
recovered by the implemented Phase~3 procedure.

The GPT-5.4 results illustrate both sources of omission.
Under No Phase~1, all final reference-edge omissions were
already present after Phase~2.
Under Full on Sachs, structural reconciliation introduced
additional losses after local judgment.
These findings indicate that final-graph evaluation alone
does not distinguish insufficient early retention from
subsequent loss of retained reference edges.
They do not imply that reconciliation should be omitted:
the intermediate graph may contain bidirectional conflicts
or directed cycles.
Rather, structural consistency and reference-edge preservation
should be assessed separately when evaluating an
omission-oriented pipeline.
The present diagnostics localize additional losses to
Phase~3 as a whole, not to a particular reconciliation substep.

\paragraph{Roles of the Phase~1 information components.}
The consistent Full--No Phase~1 result should not be
interpreted as a monotonic benefit from supplying more
pre-specified information.
Full did not uniformly outperform the partial conditions;
for example, Identity only achieved higher Recall and F1
than Full on Sachs with GPT-5.4.
The relative ordering of Identity only and Pattern only
also differed across the evaluated backbone settings.
These observations distinguish making a third-variable
hypothesis available from specifying its structural
interpretation in advance, without establishing that
either information component has a fixed role in controlling
retention or selectivity.
The GPT-4o Sachs comparisons require additional caution
because cycle resolution timed out in all Pattern-only
runs and in one Identity-only run, leading to different
reconciliation outcomes across conditions.

\paragraph{Interpretation and practical scope.}
Taken together, the results support a design-level account:
explicit pre-organization was associated with higher Recall
and F1 relative to the implemented No Phase~1 condition,
while the comparative advantage over existing methods and
the ordering of partial-information conditions varied
across settings.
These are descriptive findings from three runs per condition.
Moreover, the backbone comparisons also involve different
generation configurations, so their differences cannot be
attributed exclusively to the underlying model.

The practical rationale for omission control is strongest
when later processing filters existing candidates rather
than generates new ones.
In that setting, retained non-reference edges may still
be removed, whereas omitted reference edges are unavailable
to the filtering procedure.
However, higher retention and F1 do not establish better
downstream decisions or justify the additional inference
cost by themselves.
The study evaluates agreement with adopted reference graphs,
not independently verified causal truth, and does not
evaluate downstream review or filtering outcomes.
The value of the resulting candidate graphs therefore
depends on the costs of omissions, false positives, and
subsequent validation.
\section{Conclusion}

We introduced MaSCoD, an omission-oriented framework
that explicitly organizes third-variable hypotheses and
local structural patterns before utilizing them in
direct-edge judgment.

Across all six dataset--backbone settings, Full achieved
higher mean Recall and F1 than No Phase~1, while also
increasing FPR.
This consistent comparison with the implemented
no-pre-organization condition did not translate into
uniform superiority over existing methods:
a retention advantage over all evaluated baselines
was observed on DWD with GPT-5.4 and on Sachs with GPT-4o.
Moreover, Full did not uniformly outperform the
partial-information conditions.

The GPT-5.4 stage-wise analysis distinguished omissions
already present after local judgment from additional
losses during reconciliation.
Together, these findings support structural pre-organization
as an explicit design and evaluation target for omission
control.
They motivate evaluating how structural hypotheses are
constructed jointly with how they are used in judgment,
while separately assessing whether retained edges survive
graph-level reconciliation.

\section*{Limitations}
\label{sec:limitations}

\paragraph{Interpretation of the Phase~1 ablations.}
The four ablation conditions differ not only in the amount of
structural information supplied to Phase~2, but also in which parts of
the structural hypothesis must be inferred within the downstream
prompt.
In \textsc{Full}, both third-variable identity and structural-pattern
information are supplied; in the partial conditions, one of these is
withheld; and in \textsc{No Phase 1}, both must be hypothesized during
Phase~2.
The observed differences therefore should not be interpreted as
isolated causal effects of Phase~1 or of either information component.
Rather, they characterize differences among the implemented design
conditions, which jointly vary the supplied information and the
reasoning required from the model.

\paragraph{Evaluation scope and statistical uncertainty.}
The evaluation covers only three datasets with 5, 6, and 11 variables.
These benchmarks differ simultaneously in domain, graph size, density,
and structural characteristics, making it difficult to isolate which
dataset property accounts for a particular retention--selectivity
profile.
The results therefore should not be generalized directly to
substantially larger graphs or to domains not represented by
Auto-MPG, DWD, and Sachs.

Each LLM-based method and ablation condition was evaluated over three
runs per dataset.
We report descriptive means and sample standard deviations but do not
conduct formal significance tests or estimate confidence intervals.
Three repetitions remain limited for characterizing the stochastic
variability of LLM-based graph generation, particularly when the
observed differences between conditions are small.
The reported comparisons should therefore be interpreted
descriptively rather than as precise estimates of expected performance
differences.

\paragraph{Reference-graph and reconciliation assumptions.}
Performance is measured against reference graphs adopted from prior
work.
These graphs depend on domain assumptions and evaluation conventions
and should not be interpreted as unique or definitive representations
of the underlying causal structures.
The reported metrics therefore measure agreement with the adopted
reference graphs rather than recovery of independently verified causal
truth.

Phase~3 additionally relies on BIC under a linear-Gaussian structural
equation scoring assumption.
We do not evaluate robustness to alternative statistical scoring rules
or violations of this assumption.
Moreover, when cycle reconciliation does not complete, the graph after
bidirectional-edge resolution is retained and may still contain
directed cycles.
This fallback occurred in the GPT-4o Sachs partial ablations: cycle
resolution did not complete in all three \textsc{Pattern only} runs
and in one of the three \textsc{Identity only} runs.
The component-level Sachs results under GPT-4o should therefore be
interpreted with particular caution because the adopted outputs were
not produced under identical reconciliation outcomes across
conditions.

\paragraph{Model, decoding, and persona dependence.}
We evaluate two proprietary LLM backbones, GPT-5.4 and GPT-4o.
Although this provides a direct check of whether observed tendencies
persist across two models, it does not establish backbone invariance.
The two backbones were also evaluated under different generation
configurations: GPT-5.4 used temperature 1.0 and a maximum output
length of 1,024 tokens, whereas GPT-4o used temperature 0.2 and a
maximum output length of 512 tokens.
Consequently, differences between the two settings cannot be
attributed exclusively to the backbone itself; decoding and service
configuration may also contribute to the observed performance
profiles.
Evaluation across additional model families under more closely matched
generation settings is required to determine the generality of the
observed effects.

In addition, the three personas are instantiated using the same
underlying language model.
Their judgments are therefore not statistically independent expert
samples, and majority agreement should not be interpreted as
independent expert consensus.
The persona roles and domain specializations were manually specified,
and alternative role definitions or numbers of personas may produce
different outputs.

\paragraph{Practical utility, cost, and component attribution.}
MaSCoD requires substantially more LLM calls than a one-shot pairwise
baseline because it combines candidate generation, persona-based
evaluation, candidate selection, iterative direct-edge judgment, and
graph-level reconciliation.
This limits its applicability when latency, monetary cost, energy
consumption, or large-scale graph discovery is a primary concern.

The present experiments evaluate the candidate graphs directly against
reference graphs but do not evaluate subsequent expert review,
statistical testing, graph pruning, or application-specific decision
outcomes.
Consequently, higher edge retention does not by itself establish that
the resulting candidate graph leads to better downstream decisions.
A retention-oriented operating point may be useful when reliable
downstream filtering is available, but may be undesirable when false
positives are particularly costly.

Finally, the ablations vary the information supplied from Phase~1
but do not separately identify the contributions of the three-persona
design, the NGT-inspired candidate-organization procedure, the
Delphi-inspired judgment procedure, or the specific aggregation rules.
The reported results therefore characterize the behavior of the
implemented pipeline rather than the independent contribution of every
procedural component.
\section*{Ethical Considerations}

MaSCoD is intended for research-oriented candidate-graph generation
rather than as a substitute for domain expertise or interventional
validation. Incorrectly retained or omitted edges could lead to
misleading causal interpretations, particularly in high-stakes
applications. We therefore recommend treating the adopted graph as a
set of structural hypotheses for subsequent expert review and
statistical or experimental validation. The experiments in this study
do not involve human annotation or a newly conducted human-participant
study.

\bibliography{bibliography}

\begin{appendices}
\appendix

\section{Dataset Details}
\label{app:dataset-details}

\paragraph{LLM input.}
For the LLM-based procedures, the observational data were not provided
directly to the model.
Instead, the model received the variable names and the natural-language
descriptions listed below.
The descriptions were written to provide the semantic information
required to interpret each variable, including what it measures and how
larger or smaller values should generally be understood, without
explicitly stating the target causal directions.
Nevertheless, because such semantic descriptions may themselves inform
causal judgment, they should not be regarded as strictly causally
neutral inputs.

The wording below reproduces the descriptions used in the experiments.
We retain these descriptions without substantive revision because
changing them would alter the experimental input.
The observational datasets, preprocessing procedure, sample sizes, and
reference-graph configuration are described in
Section~\ref{sec:experimental-setup}.

\subsection{Auto-MPG}
\label{app:auto-mpg-details}

Auto-MPG is a benchmark dataset containing vehicle specifications and
fuel-efficiency measurements \cite{quinlan1993autompg}.
We use the five continuous variables
\texttt{mpg}, \texttt{displacement}, \texttt{horsepower},
\texttt{weight}, and \texttt{acceleration};
the discrete \texttt{Cylinders} variable is excluded.

\begin{table*}[t]
    \centering
    \footnotesize
    \setlength{\tabcolsep}{5pt}
    \renewcommand{\arraystretch}{1.12}
    \begin{tabularx}{\textwidth}{@{}>{\ttfamily}lX@{}}
        \toprule
        \normalfont Variable & Description \\
        \midrule
        mpg &
        Fuel efficiency (miles per gallon).
        A performance indicator that reflects overall driving
        efficiency and vehicle specifications.
        Larger values indicate better fuel efficiency; smaller values
        indicate poorer fuel efficiency.
        \\

        displacement &
        Engine displacement (cubic inches).
        A basic parameter representing total cylinder volume.
        Larger values generally indicate greater torque and fuel
        consumption; smaller values indicate a smaller engine and a
        tendency toward efficiency-oriented design.
        \\

        horsepower &
        Total horsepower (dynamometer rating).
        Represents maximum engine output.
        Larger values indicate higher output; smaller values indicate
        lower output.
        \\

        weight &
        Vehicle weight (pounds).
        The weight of the vehicle body including occupants and cargo in
        typical use.
        Larger values indicate a heavier vehicle; smaller values
        indicate a lighter vehicle and typically better maneuverability.
        \\

        acceleration &
        0--60 mph acceleration time (seconds).
        A dynamic indicator of responsiveness from standstill.
        Larger values indicate slower acceleration; smaller values
        indicate quicker acceleration.
        \\
        \bottomrule
    \end{tabularx}
    \caption{Variable descriptions supplied to the LLM for Auto-MPG.}
    \label{tab:auto-mpg-descriptions}
\end{table*}

\subsection{DWD}
\label{app:dwd-details}

The DWD dataset contains geographic, topographic, and meteorological
variables derived from observations published by the Deutscher
Wetterdienst \cite{mooij2016distinguishing}.
We use the six variables \texttt{Latitude}, \texttt{Longitude},
\texttt{Altitude}, \texttt{Precipitation}, \texttt{Sunshine}, and
\texttt{Temperature}.

\begin{table*}[t]
    \centering
    \footnotesize
    \setlength{\tabcolsep}{5pt}
    \renewcommand{\arraystretch}{1.12}
    \begin{tabularx}{\textwidth}{@{}>{\ttfamily}lX@{}}
        \toprule
        \normalfont Variable & Description \\
        \midrule
        Latitude &
        Latitude (decimal degrees).
        A coordinate indicating north--south position on the Earth.
        Larger values indicate a more northerly location; smaller
        values indicate a more southerly location.
        \\

        Longitude &
        Longitude (decimal degrees).
        A coordinate indicating east--west position on the Earth.
        Larger values indicate a location farther east; smaller values
        indicate a location farther west.
        \\

        Altitude &
        Altitude (meters).
        A topographic indicator representing the elevation of the
        observation site.
        Larger values indicate higher elevation; smaller values
        indicate lower elevation or proximity to sea level.
        \\

        Precipitation &
        Precipitation amount (mm/day).
        A daily cumulative measure of rain or snow converted into water
        equivalent.
        Larger values indicate wetter conditions; smaller values
        indicate drier conditions.
        \\

        Sunshine &
        Sunshine duration.
        Indicates the effective amount of daily sunshine.
        Larger values indicate sunnier conditions and longer sunshine
        duration; smaller values indicate cloudier conditions and less
        sunshine.
        \\

        Temperature &
        Mean air temperature ($^\circ$C/day).
        Represents the 24-hour average temperature at the observation
        site.
        Larger values indicate warmer conditions; smaller values
        indicate colder conditions.
        \\
        \bottomrule
    \end{tabularx}
    \caption{Variable descriptions supplied to the LLM for DWD.}
    \label{tab:dwd-descriptions}
\end{table*}

\subsection{Sachs}
\label{app:sachs-details}

The Sachs dataset contains continuous-valued measurements of protein
phosphorylation and intracellular signaling
\citep{sachs2005causal}.
We use the 11 variables
\texttt{raf}, \texttt{mek}, \texttt{plc}, \texttt{pip2},
\texttt{pip3}, \texttt{erk}, \texttt{akt}, \texttt{pka},
\texttt{pkc}, \texttt{p38}, and \texttt{jnk}.
Because most variable names are abbreviations, the descriptions identify
the corresponding molecule or activity state while avoiding explicit
statements of the evaluated edge directions.

\begin{table*}[t]
    \centering
    \footnotesize
    \setlength{\tabcolsep}{5pt}
    \renewcommand{\arraystretch}{1.12}
    \begin{tabularx}{\textwidth}{@{}>{\ttfamily}lX@{}}
        \toprule
        \normalfont Variable & Description \\
        \midrule
        raf &
        RAF kinase activity.
        Refers to a family of serine/threonine kinases upstream in the
        MAPK cascade.
        Larger values indicate higher activity; smaller values indicate
        lower activity.
        \\

        mek &
        MEK phosphorylation level.
        Represents the state of a dual-specificity kinase in the middle
        of the MAPK cascade.
        Larger values indicate greater phosphorylation; smaller values
        indicate a state closer to non-phosphorylated.
        \\

        plc &
        Phospholipase C activity.
        An indicator of membrane-associated enzymes that hydrolyze
        PIP2.
        Larger values indicate higher activity; smaller values indicate
        lower activity.
        \\

        pip2 &
        PIP2 abundance.
        Represents the amount of phosphoinositide in the inner leaflet
        of the cell membrane.
        Larger values indicate a larger pool; smaller values indicate a
        smaller pool.
        \\

        pip3 &
        PIP3 abundance.
        Represents the amount of phosphoinositide generated by PI3K.
        Larger values indicate greater abundance; smaller values
        indicate lower abundance.
        \\

        erk &
        ERK phosphorylation level.
        Represents the state of a downstream kinase in the MAPK
        cascade.
        Larger values indicate greater phosphorylation; smaller values
        indicate lower activity.
        \\

        akt &
        AKT kinase activity.
        Represents the output of a PIP3-dependent serine/threonine
        kinase.
        Larger values indicate higher activity; smaller values indicate
        lower activity.
        \\

        pka &
        PKA activity.
        Represents the output of a cAMP-dependent serine/threonine
        kinase.
        Larger values indicate higher activity; smaller values indicate
        lower activity.
        \\

        pkc &
        PKC activity.
        An indicator of a family of serine/threonine kinases responsive
        to DAG and Ca$^{2+}$.
        Larger values indicate higher activity; smaller values indicate
        lower activity.
        \\

        p38 &
        p38 MAPK phosphorylation level.
        Represents the state of a stress-responsive MAP kinase.
        Larger values indicate greater phosphorylation; smaller values
        indicate lower activity.
        \\

        jnk &
        JNK kinase activity.
        Represents the output of a stress-responsive MAP kinase.
        Larger values indicate higher activity; smaller values indicate
        lower activity.
        \\
        \bottomrule
    \end{tabularx}
    \caption{Variable descriptions supplied to the LLM for Sachs.}
    \label{tab:sachs-descriptions}
\end{table*}

\section{Reference Graphs}
\label{app:reference-graphs}

This appendix reports the reference graphs used for evaluation.
For Auto-MPG, DWD, and Sachs, we follow the evaluation configuration
adopted in prior work \cite{Takayama2025}.
These graphs are used as benchmark reference structures and should not
be interpreted as uniquely verified causal graphs for the corresponding
domains.
Accordingly, the reported metrics quantify agreement with the adopted
reference graphs.

\subsection{Auto-MPG Reference Graph}

The Auto-MPG reference graph contains five directed edges.
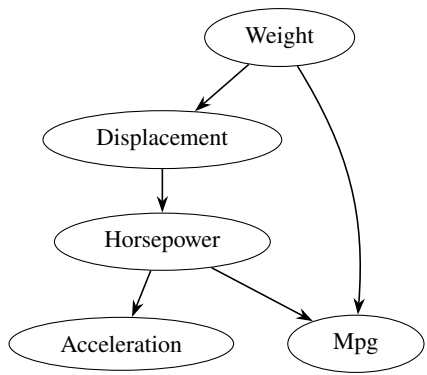
\begin{figure}[tb]
  \centering
  \begin{tikzpicture}[x=1.55cm,y=1.35cm]
    \node[causalnode] (weight) at (0,3)
      {Weight};

    \node[causalnode] (disp) at (-1,2)
      {Displacement};

    \node[causalnode] (hp) at (-1,1)
      {Horsepower};

    \node[causalnode] (acc) at (-1.35,0)
      {Acceleration};

    \node[causalnode] (mpg) at (0.65,0)
      {Mpg};

    \draw[causal] (weight) -- (disp);
    \draw[causal] (disp) -- (hp);
    \draw[causal] (hp) -- (acc);
    \draw[causal] (hp) -- (mpg);
    \draw[causal] (weight)
      to[bend left=18] (mpg);
  \end{tikzpicture}

  \caption{
    Reference graph adopted for Auto-MPG data.}
  \label{fig:auto-mpg-ground-truth}
\end{figure}

\subsection{DWD Reference Graph}

The DWD reference graph contains six directed edges.
\begin{figure}[htbp]
  \centering
  \resizebox{0.78\linewidth}{!}{%
  \begin{tikzpicture}[x=1.65cm,y=1.35cm]
    \node[causalnode] (lat)  at (-2.5,1.5) {Latitude};
    \node[causalnode] (lon)  at (0,1.5) {Longitude};
    \node[causalnode] (alt)  at (2.5,1.5) {Altitude};

    \node[causalnode] (temp) at (-1.8,0) {Temperature};
    \node[causalnode] (prec) at (0.75,0) {Precipitation};
    \node[causalnode] (sun)  at (3.05,0) {Sunshine};

    \draw[causal] (lat) -- (temp);
    \draw[causal] (lon) -- (temp);
    \draw[causal] (alt) -- (temp);
    \draw[causal] (lon) -- (prec);
    \draw[causal] (alt) -- (prec);
    \draw[causal] (alt) -- (sun);
  \end{tikzpicture}
  }
  \caption{Reference graph adopted for the DWD climate data.}
  \label{fig:gt-dwd}
\end{figure}
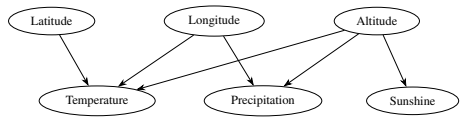
The graph is used as an evaluation reference rather than as a claim
that these six relations constitute a unique complete causal model of
the climate system.

\subsection{Sachs Reference Graph}

The Sachs reference graph contains nineteen directed edges:
\begin{figure}[htbp]
  \centering
  \resizebox{0.78\linewidth}{!}{%
  \begin{tikzpicture}[x=1.45cm,y=1.12cm]
    \node[causalnode] (pip3) at (0,6.0) {pip3};
    \node[causalnode] (plc)  at (1.35,5.1) {plc};
    \node[causalnode] (pip2) at (0.55,4.2) {pip2};
    \node[causalnode] (pkc)  at (1.35,3.15) {pkc};

    \node[causalnode] (raf)  at (0.25,2.05) {raf};
    \node[causalnode] (pka)  at (2.25,2.05) {pka};

    \node[causalnode] (mek)  at (-0.55,0.95) {mek};
    \node[causalnode] (p38)  at (3.35,0.95) {p38};
    \node[causalnode] (jnk)  at (4.75,0.95) {jnk};

    \node[causalnode] (erk)  at (-0.05,-0.25) {erk};
    \node[causalnode] (akt)  at (-0.05,-1.50) {akt};

    \draw[causal] (pip3) -- (plc);
    \draw[causal] (pip3) -- (pip2);
    \draw[causal] (plc) -- (pip2);
    \draw[causal] (plc) -- (pkc);
    \draw[causal] (pip2) -- (pkc);

    \draw[causal] (pkc) -- (raf);
    \draw[causal] (pkc) to[bend right=44] (mek);
    \draw[causal] (pkc) -- (pka);
    \draw[causal] (pkc) to[bend left=28] (p38);
    \draw[causal] (pkc) to[bend left=18] (jnk);

    \draw[causal] (raf) -- (mek);
    \draw[causal] (mek) -- (erk);
    \draw[causal] (erk) -- (akt);

    \draw[causal] (pka) -- (mek);
    \draw[causal] (pka) -- (erk);
    \draw[causal] (pka) to[bend left=14] (akt);
    \draw[causal] (pka) -- (p38);
    \draw[causal] (pka) -- (jnk);

    \draw[causal] (pip3) to[out=205,in=160,looseness=1.25] (akt);
  \end{tikzpicture}%
  }
  \caption{Reference graph adopted for the Sachs protein data.}
  \label{fig:gt-sachs}
\end{figure}
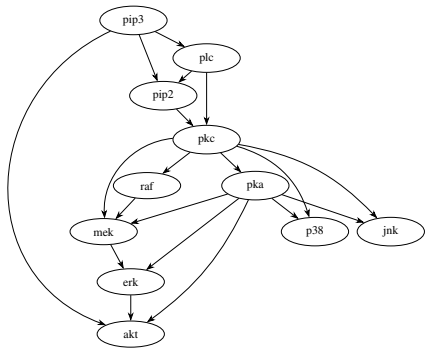

As with DWD, this graph is treated as an adopted evaluation reference
rather than as a uniquely established biological causal network.

\subsection{Evaluation Interpretation}

All Precision, Recall, F1, FPR, FNR, and NHD values in this paper are
computed relative to the reference graphs above.
The complete numerical results are reported in
Appendix~\ref{app:full-results}.
Thus, statements about retained, omitted, or non-reference edges refer
to agreement with these adopted reference structures rather than to
independently verified causal truth.

\section{Persona Specialization Definitions}
\label{app:personas}

We use three manually specified personas for each dataset to introduce
different domain-relevant perspectives into candidate generation and
causal judgment.
Within each stage, the personas receive the same task definition and
pair-specific information; only their role framing and specialization
differ.
No persona is given the reference graph or the correct label for any
variable pair.
The persona mechanism is therefore intended to expose different
considerations within the prompt, rather than to encode the evaluated
causal structure.

All personas are instantiated using the same underlying language model.
Their outputs should consequently not be interpreted as statistically
independent expert judgments, and majority agreement does not constitute
independent human-expert consensus.
The number of personas and their specializations were not optimized;
the configurations in Tables~\ref{tab:personas-auto},
\ref{tab:personas-dwd}, and~\ref{tab:personas-sachs} were fixed
throughout the experiments.
The tables reproduce the persona specifications used in the prompts.

\begin{table*}[t]
    \centering
    \footnotesize
    \setlength{\tabcolsep}{5pt}
    \renewcommand{\arraystretch}{1.10}
    \begin{tabularx}{\textwidth}{
        @{}
        >{\raggedright\arraybackslash}p{0.19\textwidth}
        >{\raggedright\arraybackslash}X
        >{\raggedright\arraybackslash}X
        @{}
    }
        \toprule
        Role & Specialty description & Primary perspective \\
        \midrule

        Statistician &
        Classical regression, causal inference, and diagnosis of
        multicollinearity and confounding. &
        Organizing variable dependencies, assessing possible
        confounding, and interpreting statistical relationships.
        \\

        Automotive engineer &
        Powertrain dynamics, combustion efficiency, drivetrain design,
        and vehicle integration. &
        Vehicle mechanisms, physical laws, and mechanistic consistency
        among performance indicators.
        \\

        Machine learning engineer &
        Feature engineering, leakage detection, reproducible pipelines,
        and model monitoring. &
        Separation of variable roles, feature redundancy, and predictive
        interpretability.
        \\

        \bottomrule
    \end{tabularx}
    \caption{Persona configurations used for Auto-MPG.}
    \label{tab:personas-auto}
\end{table*}

\begin{table*}[t]
    \centering
    \footnotesize
    \setlength{\tabcolsep}{5pt}
    \renewcommand{\arraystretch}{1.10}
    \begin{tabularx}{\textwidth}{
        @{}
        >{\raggedright\arraybackslash}p{0.19\textwidth}
        >{\raggedright\arraybackslash}X
        >{\raggedright\arraybackslash}X
        @{}
    }
        \toprule
        Role & Specialty description & Primary perspective \\
        \midrule

        Lead climatologist &
        Observation quality control, climate variability, and extreme
        events. &
        Meaning of meteorological observations, physical consistency,
        and climatological interpretation.
        \\

        Geospatial analyst &
        Spatial referencing, coordinate-system consistency,
        terrain-aware interpolation, and coverage diagnostics. &
        Interpretation of geographic attributes such as latitude,
        longitude, and altitude, and the influence of spatial
        configuration.
        \\

        Causal statistics lead &
        Time-series decomposition, generalized additive models, and
        causal sensitivity analysis for climate factors. &
        Organizing structural dependence, assessing possible
        confounding, and interpreting statistical models.
        \\

        \bottomrule
    \end{tabularx}
    \caption{Persona configurations used for DWD.}
    \label{tab:personas-dwd}
\end{table*}

\begin{table*}[t]
    \centering
    \footnotesize
    \setlength{\tabcolsep}{5pt}
    \renewcommand{\arraystretch}{1.10}
    \begin{tabularx}{\textwidth}{
        @{}
        >{\raggedright\arraybackslash}p{0.19\textwidth}
        >{\raggedright\arraybackslash}X
        >{\raggedright\arraybackslash}X
        @{}
    }
        \toprule
        Role & Specialty description & Primary perspective \\
        \midrule

        Systems biologist &
        Graphical causal models, biochemical network inference, and
        perturbation analysis. &
        Overall network structure, pathway dependencies, and
        consistency of structural hypotheses.
        \\

        Molecular signaling engineer &
        Kinase cascades, phosphoproteomics, and feedback control in
        T-cell signaling assays. &
        Local mechanisms, molecular reactions, and ordering relations
        along signaling pathways.
        \\

        Experimental scientist &
        Flow cytometry, multiplex assays, and quality control for
        repeated experiments. &
        Meaning of measured quantities, experimental conditions, and
        interpretation of observations.
        \\

        \bottomrule
    \end{tabularx}
    \caption{Persona configurations used for Sachs.}
    \label{tab:personas-sachs}
\end{table*}

\section{Overall Workflow Pseudocode}

This appendix presents the overall workflow of the proposed method in algorithmic form. In the main text, the method is described as a staged procedure consisting of Phase 1 through Phase 3. However, from the textual description alone, it may not be immediately clear at which unit the inputs are processed and how the output of each phase is passed to the next phase. This appendix therefore makes explicit the overall dependency structure from the input variable set, natural-language descriptions, observational data, and persona set to the final adopted graph.

The proposed method consists of four major steps. First, for each ordered variable pair, Phase 1 selects candidate third variables and structural patterns. Second, conditioned on the candidate set obtained in Phase 1, Phase 2 determines whether a direct effect remains for the ordered pair. Third, when both directions survive for the same unordered variable
pair, Phase~3 performs a binary comparison between the two directions.
Each of the three personas must select either \(X\rightarrow Y\) or
\(Y\rightarrow X\), and the direction receiving at least two votes is
retained. Finally, if directed cycles remain after bidirectional-edge resolution, a BIC-based post-hoc cycle-removal procedure is applied using the observational data, yielding the final adopted graph.

The key distinction is that Phases 1 and 2 are executed at the level of ordered variable pairs, whereas Phase 3 and the subsequent cycle-removal step operate on the graph obtained by aggregating the Phase 2 outputs. In other words, the first part of the procedure accumulates local judgments, and the later part reconciles those judgments into a globally coherent graph. Making this difference in processing units explicit is important for understanding the proposed method.

The pseudocode in this appendix is not intended to restate the full internal procedure of each phase. The details of candidate generation and aggregation in Phase~1 are provided in Appendix~F, the detailed judgment procedure in Phase~2 is provided in Appendix~G, and the Phase~3 procedures, including bidirectional-edge resolution and cycle removal, are provided in
Appendix~I. The role of this appendix is to connect these component algorithms at a higher level and to clarify the execution order and data flow of the full method. 

\begin{table*}[t]
\centering
\small
\refstepcounter{algorithm}
\label{alg:overall-workflow}
\begin{tabularx}{\textwidth}{@{}r@{\quad}X@{}}
\toprule
\multicolumn{2}{@{}l}{\textbf{Algorithm \thealgorithm} Overall Workflow of the Proposed Method} \\
\midrule
1: & \textbf{Input:} variable set \(V=\{X_1,\ldots,X_n\}\), natural-language descriptions for all variables in \(V\), observational data \(D\), and persona set \(P=\{p_1,p_2,p_3\}\). \\
2: & \textbf{Output:} adopted directed graph \(G^{*}\), which is acyclic if no directed cycle remains after bidirectional-edge resolution or if post-hoc cycle removal completes. \\
3: & Enumerate all ordered variable pairs \((X,Y)\) such that \(X,Y\in V\) and \(X\neq Y\). \\
4: & Initialize \(L \leftarrow \emptyset\). \\
5: & \textbf{for all} ordered pairs \((X,Y)\) \textbf{do} \\
6: & \quad Run Phase 1 for \((X,Y)\), using the variable descriptions and persona set \(P\), and obtain the selected candidate set \(C_{XY}\). \\
7: & \quad Run Phase 2 for \((X,Y)\), conditioned on \(C_{XY}\), and obtain the final pairwise label \(\ell_{XY}\). \\
8: & \quad Add \(\ell_{XY}\) to \(L\). \\
9: & \textbf{end for} \\
10: & Map all pairwise labels in \(L\) into directed edge candidates and construct an intermediate directed graph \(G\). \\
11: & Specifically, if \(\ell_{XY}\) is mapped to the presence of a direct effect from \(X\) to \(Y\), add \(X\to Y\) to \(G\); otherwise, do not add \(X\to Y\). \\
12: & \textbf{if} \(G\) contains both \(X\to Y\) and \(Y\to X\) for any unordered variable pair \textbf{then} \\
13: & \quad Run Phase 3 bidirectional-edge resolution. \\
14: & \quad Update \(G\) by retaining the selected direction according to the Phase 3 decision rule. \\
15: & \textbf{end if} \\
16: & Store the graph obtained after bidirectional-edge resolution as \(G_{\mathrm{bi}}\). \\
17: & \textbf{if} \(G_{\mathrm{bi}}\) contains one or more directed cycles \textbf{then} \\
18: & \quad Attempt post-hoc cycle removal based on BIC using observational data \(D\). \\
19: & \quad \textbf{if} cycle removal completes and produces an acyclic output \textbf{then} \\
20: & \qquad Update \(G\) to the selected acyclic graph. \\
21: & \quad \textbf{else} \\
22: & \qquad Set \(G \leftarrow G_{\mathrm{bi}}\). \\
23: & \quad \textbf{end if} \\
24: & \textbf{else} \\
25: & \quad Set \(G \leftarrow G_{\mathrm{bi}}\). \\
26: & \textbf{end if} \\
27: & \(G^{*} \leftarrow G\). \\
28: & \textbf{return} \(G^{*}\). \\
\bottomrule
\end{tabularx}
\end{table*}

As shown in Algorithm D.1, the proposed method first performs local judgments for all ordered variable pairs and then integrates those judgments into an intermediate graph. It then processes two types of structural inconsistency in sequence: bidirectional edges and directed cycles. Bidirectional-edge resolution addresses conflicts that arise when
opposite local judgments both survive for the same unordered pair.
For each conflict, three personas independently make a binary choice
between the two directions, and the direction receiving at least two
valid votes is retained.
Cycle removal addresses cases in which the graph still violates the
DAG constraint after bidirectional-edge resolution. Because these two inconsistencies have different roles, they are handled in separate stages.

The final cycle-removal step is not applied to every pairwise judgment. It is invoked only when directed cycles remain after Phase 3. This distinction is important because the core of the proposed method lies in LLM-based local judgment, while the BIC-based step is a minimal statistical reconciliation procedure used only when necessary to obtain a final adopted graph. 

The phrase ``map all pairwise labels into directed edge candidates'' in Algorithm D.1 summarizes the conversion from Phase 2 labels to graph edges. In the implementation, labels containing \texttt{Direct} are mapped to the presence of the corresponding directed edge, whereas labels without \texttt{Direct} and the \texttt{Invalid} label are mapped to the absence of that edge. The intermediate graph is then constructed by aggregating these edge-level decisions over all ordered pairs. 

The purpose of this algorithm is to clarify the overall execution order of the proposed method. It does not expand the prompt content, majority-vote rules, candidate-scoring procedure, or BIC computation used in each component. Those details are separated into the corresponding appendices so that readers can first understand the global workflow and then inspect the implementation details of each phase as needed. 

\paragraph{Software environment.}
The experiments were implemented in Python using
\texttt{causal-learn} 0.1.3.6, NumPy 1.26.4, pandas 2.3.3,
scikit-learn 1.7.1, and SciPy 1.16.1.
Semantic deduplication used
\texttt{sentence-transformers/all-MiniLM-L6-v2} with the
\texttt{community\_detection} procedure and the settings described in
Appendix~F.

\section{Phase 1 Candidate-Generation Prompt}
\label{app:phase1-prompt}

This appendix reproduces the prompt used to generate candidate third
variables in Phase~1.
The experiments used the original Japanese prompt shown below.
The English version is provided only for readability and was not used
as a separate experimental condition.
For each ordered variable pair \((X,Y)\), the prompt asks each persona
to propose third variables \(Z\) and assign each candidate a Chain,
Fork, or Collider pattern.
This stage generates auxiliary structural hypotheses; it does not
determine whether the direct edge \(X\rightarrow Y\) is retained.
Candidate merging, scoring, and final Phase~1 selection are described
in Appendix~\ref{app:phase1-algorithm}.

\paragraph{Prompt E.1: Original Japanese prompt.}

\begin{quote}
\small

\textbf{【タスク】}

変数ペア \(X\rightarrow Y\) の関係を理解するために考慮すべき
第三要因 \(Z\) を列挙してください。

\textbf{【第三要因の構造】}

第三要因は次の3つの構造を意識して挙げてください。

\begin{enumerate}
    \item \textbf{媒介（chain）}

    \(X\) が \(Z\) を経由して \(Y\) に影響する。

    例：気温 \(\rightarrow\) 需要 \(\rightarrow\) 売上

    \item \textbf{共通原因（fork）}

    \(Z\) が \(X\) と \(Y\) の両方に影響する。

    例：景気 \(\rightarrow\) 消費と株価

    \item \textbf{共通結果（collider）}

    \(X\) と \(Y\) が \(Z\) に影響する。

    例：雨量と交通量 \(\rightarrow\) 渋滞
\end{enumerate}

\textbf{【候補変数の条件】}

各候補 \(Z\) について以下を満たしてください。

\begin{itemize}
    \item
    \(X\)--\(Z\) および \(Z\)--\(Y\) の関係が、ドメイン知識・
    物理法則・時間順序と整合するかを確認する。

    \item
    整合する理由を1文で説明し、法則名・メカニズム・式の
    いずれかを最低1つ挙げる。

    \item
    事前推定が提示される場合は参考にしてよい。
    ただし可能性ベースの情報のため、矛盾の可能性だけで
    機械的に除外しない。

    \item
    観測変数以外（未観測変数）も提示可能。
    ただし、同じ役割の観測変数が
    \texttt{VARS} にある場合は観測変数を優先する。

    \item
    \(Z\) は \(X\) と \(Y\) の両方と因果的または機構的な
    関係を持つ可能性があるものに限定する。
\end{itemize}

\textbf{【対象変数】}

\texttt{<PAIR>}%
\texttt{<X>} \(\rightarrow\) \texttt{<Y>}%
\texttt{</PAIR>}

\texttt{<VARS>}

\texttt{<変数集合の説明文（JSON形式）>}

\texttt{</VARS>}

\textbf{【出力形式】}

JSON配列のみを出力してください。

\begin{verbatim}
[
  {
    "var": "...",
    "pattern": "chain|fork|collider",
    "reason": "..."
  }
]
\end{verbatim}

\end{quote}

\paragraph{Prompt E.2: English translation.}

\begin{quote}
\small

\textbf{[Task]}

List candidate third variables \(Z\) that should be considered when
interpreting the relation \(X\rightarrow Y\).

\textbf{[Third-variable structures]}

Consider the following three structural patterns when proposing third
variables.

\begin{enumerate}
    \item \textbf{Chain}

    \(X\) affects \(Y\) through \(Z\).

    Example:
    temperature \(\rightarrow\) demand \(\rightarrow\) sales.

    \item \textbf{Fork}

    \(Z\) affects both \(X\) and \(Y\).

    Example:
    economic conditions \(\rightarrow\) consumption and stock prices.

    \item \textbf{Collider}

    \(X\) and \(Y\) both affect \(Z\).

    Example:
    rainfall and traffic volume \(\rightarrow\) congestion.
\end{enumerate}

\textbf{[Conditions for candidate variables]}

Each proposed candidate \(Z\) must satisfy the following conditions.

\begin{itemize}
    \item
    Check whether the \(X\)--\(Z\) and \(Z\)--\(Y\) relations are
    consistent with domain knowledge, physical laws, and temporal
    ordering.

    \item
    Explain the reason for this consistency in one sentence and mention
    at least one relevant law, mechanism, or equation.

    \item
    If prior estimates are provided, they may be used as supporting
    information.
    However, because they represent possibilities rather than
    established facts, do not mechanically exclude a candidate solely
    because of a possible inconsistency.

    \item
    Variables not included in the observed variable set may also be
    proposed.
    However, when an observed variable in \texttt{VARS} can play the
    same role, prioritize the observed variable.

    \item
    Restrict \(Z\) to variables that may have a causal or mechanistic
    relation with both \(X\) and \(Y\).
\end{itemize}

\textbf{[Target variables]}

\texttt{<PAIR>}

\texttt{<X> -> <Y>}

\texttt{</PAIR>}

\texttt{<VARS>}

\texttt{<Variable descriptions in JSON format>}

\texttt{</VARS>}

\textbf{[Output format]}

Output only a JSON array.

\begin{verbatim}
[
  {
    "var": "...",
    "pattern": "chain|fork|collider",
    "reason": "..."
  }
]
\end{verbatim}

\end{quote}

The generated candidates are not passed directly to Phase~2.
They are first pooled across personas, semantically deduplicated,
assigned a single structural pattern, scored, and subjected to a
binary acceptance vote as described in
Appendix~\ref{app:phase1-algorithm}.

\section{Phase 1 Candidate Aggregation and Selection}
\label{app:phase1-algorithm}

Phase~1 converts the third-variable hypotheses generated for each
ordered pair \((X,Y)\) into a consolidated candidate set
\[
    C_{XY}
    =
    \{(z_m,r_m)\}_{m=1}^{M_{XY}},
\]
where \(z_m\) denotes a selected third-variable candidate and
\(r_m\in
\{\textsc{Chain},\textsc{Fork},\textsc{Collider}\}\)
denotes the structural pattern assigned to that candidate.
The candidates in \(C_{XY}\) are auxiliary structural hypotheses for
Phase~2 and are not inserted directly into the final graph.

\paragraph{Candidate elicitation and refinement.}
For each ordered pair, the three persona processes first generate
third-variable candidates using the prompt in
Appendix~\ref{app:phase1-prompt}.
The initial candidate-generation calls do not receive the other
personas' candidate lists.
The implementation retains at most 12 initially generated candidates
per persona.

After initial elicitation, the procedure performs reflection,
cross-persona question answering, and a self-check before candidate
consolidation.
The implementation may either explicitly regenerate revised candidate
opinions or synthesize them from the self-check output, depending on
the configured execution path.
The initial and revised or synthesized candidate records are then
pooled for semantic deduplication.
These intermediate exchanges are used to refine the candidate pool;
they do not themselves determine whether a candidate is retained in
\(C_{XY}\).

\paragraph{Semantic deduplication and pattern consolidation.}
Semantic deduplication is performed on the candidate-name strings only.
Each candidate name is embedded using
\texttt{sentence-transformers/all-MiniLM-L6-v2}.
The resulting embeddings are clustered using the
\texttt{community\_detection} procedure from Sentence Transformers
with a cosine-similarity threshold of
\[
    \tau_{\mathrm{dedup}} = 0.86
\]
and \texttt{min\_community\_size}=1.
Natural-language candidate descriptions and reasons are not included
in the embedding used for this deduplication step.

Within each resulting cluster, the shortest candidate-name string is
used as the representative form.
If multiple strings have the same minimum length, the candidate that
appears earlier in the input record order is selected.
Each cluster is assigned a single structural pattern by counting the
\textsc{Chain}, \textsc{Fork}, and \textsc{Collider} labels among its
member records and retaining the most frequent pattern.
The implementation applies no separate semantic tie-breaking procedure
when multiple patterns have the same count; the first maximum returned
under the member-record ordering is used.
Thus, this pattern assignment should be interpreted as consolidation
of the generated hypotheses rather than as a determination of the
true structural role of the third variable.

\paragraph{Candidate scoring and narrowing.}
After deduplication, each persona scores candidate importance for
interpreting the direct relation \(X\rightarrow Y\).
For candidate \(z\), let
\[
    s_{p,z}\in\{1,\ldots,10\}
\]
denote the valid score retained for persona \(p\).
Although the scoring prompt contains a textual 0-point category, the
implementation validates accepted scores on the interval
\(1\)--\(10\); invalid or out-of-range score items are subjected to a
repair procedure before aggregation.

Each persona retains at most 10 scored candidates.
If the same candidate identifier occurs more than once in a persona's
response, only that persona's highest valid score for the candidate is
retained.
The aggregate score is then
\[
    S_z
    =
    \sum_{p:\,z\text{ scored by }p} s_{p,z}.
\]
A candidate omitted from a persona's retained score list contributes
no score from that persona.
Candidates are ranked by \(S_z\), and at most the top 10 candidates are
passed to the final acceptance procedure.
The aggregate score is therefore used for prioritization rather than
as a direct acceptance threshold.

The implementation also summarizes the scoring reasons.
When the score pattern indicates remaining disagreement, additional
reflection and cross-persona question answering may be performed;
otherwise, these follow-up exchanges are skipped.
A final self-check is conducted before the acceptance vote.

\paragraph{Final candidate acceptance.}
For each candidate entering the final vote, each persona is asked to
decide whether the candidate should be considered in Phase~2 using a
binary \texttt{Yes}/\texttt{No} judgment.
The voting context includes the consolidated candidate and pattern,
its description, the scoring information and summarized reasons, and
any available follow-up and self-check information.

Invalid Yes/No records are excluded, and duplicate votes for the same
candidate from the same persona are reduced to the first valid vote.
For candidate \(z\), the implementation computes
\[
    \operatorname{YesRate}(z)
    =
    \frac{N_{\mathrm{Yes}}(z)}
         {N_{\mathrm{Yes}}(z)+N_{\mathrm{No}}(z)},
\]
using the valid votes obtained for that candidate.
The candidate is retained when
\[
    \operatorname{YesRate}(z)\geq 0.6.
\]
When all three persona votes are valid, this criterion is equivalent to
requiring at least two \texttt{Yes} votes.
If fewer than three valid votes are available, the denominator contains
only the valid Yes/No votes.

The resulting Phase~1 output is
\[
    C_{XY}
    =
    \left\{
        (z,r_z)
        \;\middle|\;
        \operatorname{YesRate}(z)\geq 0.6
    \right\}.
\]
Accordingly, Phase~1 does not establish whether a proposed third
variable is causally true.
Rather, it converts open-ended candidate generation into a consolidated
set of structural hypotheses to be examined during the Phase~2
thought experiment.

\begin{algorithm}[t]
    \caption{Phase~1 candidate aggregation and selection}
    \label{alg:phase1}
    \small
    \begin{algorithmic}[1]
        \Require
        Ordered variable pair \((X,Y)\);
        variable descriptions \(\mathbf{d}\);
        persona set
        \(\mathcal{P}=\{p_1,p_2,p_3\}\)

        \Ensure
        Selected candidate set
        \(C_{XY}\)

        \State Initialize raw candidate records
        \(\mathcal{R}_{XY}\gets\emptyset\)

        \ForAll{\(p\in\mathcal{P}\)}
            \State
            \(L_p\gets
            \Call{GenerateCandidates}
            {X,Y,\mathbf{d},p}\)
            \Comment{Appendix~\ref{app:phase1-prompt}}
            \State Retain at most 12 initial candidates
        \EndFor

        \State Perform reflection, cross-persona Q\&A, and self-check
        \State Obtain revised or synthesized candidate records
        \State Pool initial and revised records into
        \(\mathcal{R}_{XY}\)

        \State Embed only the candidate-name strings in
        \(\mathcal{R}_{XY}\) using
        \texttt{all-MiniLM-L6-v2}

        \State
        \(\mathcal{M}_{XY}\gets
        \Call{CommunityDetection}
        {\mathcal{R}_{XY},\,\tau=0.86}\)

        \ForAll{clusters \(m\in\mathcal{M}_{XY}\)}
            \State Select the shortest member name as \(z_m\)
            \State Break equal-length ties by input order
            \State Set \(r_m\) to the most frequent member-record pattern
        \EndFor

        \ForAll{\(p\in\mathcal{P}\)}
            \State Score at most 10 consolidated candidates on
            \(1\)--\(10\)
            \State For duplicate candidate IDs, retain the highest score
        \EndFor

        \ForAll{consolidated candidates \(z\)}
            \State
            \(S_z\gets
            \sum_{p:\,z\text{ scored by }p}s_{p,z}\)
        \EndFor

        \State Retain at most the 10 candidates with largest \(S_z\)
        as the voting set \(\mathcal{V}_{XY}\)

        \State Perform follow-up reflection/Q\&A when required
        \State Perform the final self-check

        \State Initialize \(C_{XY}\gets\emptyset\)

        \ForAll{\(z\in\mathcal{V}_{XY}\)}
            \ForAll{\(p\in\mathcal{P}\)}
                \State Collect a valid
                \(a_{p,z}\in\{\texttt{Yes},\texttt{No}\}\), if available
            \EndFor

            \State
            \(q_z\gets
            N_{\mathrm{Yes}}(z)/
            \bigl(
                N_{\mathrm{Yes}}(z)+N_{\mathrm{No}}(z)
            \bigr)\)

            \If{\(q_z\geq0.6\)}
                \State
                \(C_{XY}\gets
                C_{XY}\cup\{(z,r_z)\}\)
            \EndIf
        \EndFor

        \State \Return \(C_{XY}\)
    \end{algorithmic}
\end{algorithm}

\section{Full Phase 2 Thought-Experiment Prompt}
\label{app:phase2-prompt}

Phase~2 evaluates whether a direct edge \(X\rightarrow Y\) should
remain after considering the structural context produced in Phase~1.
For each retained third-variable candidate \(Z\), the method evaluates
the ordered pair under the structural pattern assigned to \(Z\).
When multiple candidate-level judgments are obtained, they are
subsequently integrated into one pair-level judgment.

The purpose of Phase~2 is not to establish that the supplied
third-variable hypothesis is itself true.
Rather, \(Z\) and its structural pattern define an auxiliary
counterfactual condition under which the persistence of a direct
\(X\rightarrow Y\) effect is examined.
The prompt therefore treats the Phase~1 output as a hypothesis that may
be rejected when it conflicts with temporal ordering, established
mechanisms, or domain knowledge.

The experiments used Japanese prompts.
English translations are provided for readability.
Pattern-specific sections for
\textsc{Chain}, \textsc{Fork}, and \textsc{Collider}, together with
the corresponding comparison conditions, are given in
Appendix~\ref{app:phase2-differences}.

\subsection{Phase 2 Workflow}
\label{app:phase2-workflow}

Let
\[
    C_{XY}
    =
    \{(z_c,r_c)\}_{c=1}^{M_{XY}}
\]
be the candidate set produced by Phase~1.
For each candidate \(c\), Phase~2 performs a candidate-level
direct-causality judgment and obtains a label \(j_c\) together with a
reason.
 The possible labels are
  \textsc{Fork},
  \textsc{Fork+Direct},
  \textsc{Chain},
  \textsc{Chain+Direct},
  \textsc{Collider},
  \textsc{Collider+Direct},
  and \textsc{Invalid}.

When a structural pattern is specified, the available labels are
restricted to the two labels associated with that pattern plus
\textsc{Invalid}.
For example, a \textsc{Chain} candidate permits
\textsc{Chain}, \textsc{Chain+Direct}, and \textsc{Invalid}.
When no candidate is predefined, all structural alternatives are
available and the model must first hypothesize a third variable and its
structural pattern.

Labels containing \textsc{Direct} indicate that a direct
\(X\rightarrow Y\) path remains after considering the structural
alternative.
Labels without \textsc{Direct} indicate that the corresponding
structural pattern explains the relation without retaining the direct
edge.
\textsc{Invalid} indicates that the direction under evaluation is not
supported under the supplied or hypothesized structural context.
It does not by itself constitute evidence for \(Y\rightarrow X\).

\begin{algorithm}[t]
    \caption{Phase~2 direct-effect judgment}
    \label{alg:phase2}
    \small
    \begin{algorithmic}[1]
        \Require
        Ordered pair \((X,Y)\);
        Phase~1 candidate set \(C_{XY}\);
        variable descriptions \(\mathbf{d}\);
        persona set \(\mathcal{P}=\{p_1,p_2,p_3\}\)

        \Ensure
        Final Phase~2 label \(\ell_{XY}\)

        \State Initialize candidate-level judgments
        \(J_{XY}\gets\emptyset\)

        \If{\(C_{XY}=\emptyset\)}
            \State Run the no-predefined-candidate judgment
            \State Obtain \((j,\rho)\)
            \State \Return \(j\)
        \EndIf

        \ForAll{\((z_c,r_c)\in C_{XY}\)}
            \State Run candidate-level judgment for
            \((X,Y,z_c,r_c)\)
            \State Obtain label \(j_c\) and reason summary \(\rho_c\)
            \State
            \(J_{XY}\gets
            J_{XY}\cup\{(z_c,j_c,\rho_c)\}\)
        \EndFor

        \If{\(|J_{XY}|=1\)}
            \State \Return the single candidate-level label
        \Else
            \State
            \(\ell_{XY}\gets
            \Call{SelectAcrossCandidates}{X,Y,J_{XY},\mathcal{P}}\)
            \State \Return \(\ell_{XY}\)
        \EndIf
    \end{algorithmic}
\end{algorithm}

\subsection{Candidate-Level Direct-Causality Judgment}
\label{app:phase2-candidate}

For each candidate \(Z\), the prompt asks whether a causal influence
from \(X\) to \(Y\) remains after the structural role of \(Z\) is taken
into account.
The judgment is based on a conceptual intervention on \(X\), rather
than requiring \(X\) to be directly manipulable in a physical
experiment.
The intervention may therefore correspond to a change in design,
condition, policy, environment, composition, or another
counterfactual modification that changes the state represented by
\(X\).

The prompt first requires a plausible mechanism specific to
\(X\rightarrow Y\) to be considered.
A variable is not rejected merely because it is difficult to manipulate
directly or appears to be a measured attribute.
\textsc{Invalid} is reserved for cases in which
\(X\rightarrow Y\) is clearly inconsistent with temporal ordering,
physical or domain constraints, or can be rationally explained only in
the reverse direction.
A mechanism supporting \(Y\rightarrow X\) may not be reused as
evidence for \(X\rightarrow Y\).

The treatment of \(Z\) depends on its structural pattern.
For \textsc{Fork} and \textsc{Chain}, the thought experiment considers
whether an effect of changing \(X\) on \(Y\) remains after conceptually
holding \(Z\) fixed.
For \textsc{Collider}, the judgment is instead made without
conditioning on \(Z\), because conditioning on a collider may induce a
spurious association.
The exact pattern-specific instructions are given in
Appendix~\ref{app:phase2-differences}.

A \textsc{Direct} label is selected only when the remaining change in
\(Y\) can be supported by a mechanism specific to
\(X\rightarrow Y\), rather than merely by a path that implicitly
changes \(Z\) or by another indirect explanation.

\subsection{Common Thought-Experiment Prompt}
\label{app:phase2-common-prompt}

The following presents the common content of the Japanese prompt.
The pattern-dependent blocks are expanded in
Appendix~\ref{app:phase2-differences}.

\paragraph{Prompt G.1: Original Japanese prompt skeleton.}

\begin{quote}
\small

あなたは
\texttt{\{expertise\}}
を専門とする
\texttt{\{role\}}
です。
あなたの専門分野の視点から、与えられた事象間の因果関係の
有無についてタスクを実施してください。

\textbf{【タスク】}

\(X\rightarrow Y\) の因果関係を評価し、第三変数 \(Z\) の
可能性も考慮したうえで、\(X\) から \(Y\) への直接効果
（direct effect）の有無を判定してください。

直接効果とは、\(Z\) を経由しない
\(X\rightarrow Y\) の因果経路による影響を指します。

\(Z\) は補助的な仮説として扱い、既知の因果メカニズム
（法則や関係式など）、一般知識・ドメイン知識、または
時間順序と明確に矛盾する場合は、その仮定に依存しすぎないで
ください。

なお、\(Y\rightarrow X\) のメカニズムを
\(X\rightarrow Y\) の根拠として用いてはいけません。

まず \(X\rightarrow Y\) の因果メカニズムを想定し、
その想定が物理法則・時間順序・ドメイン知識と明確に矛盾する、
または実質的に \(Y\rightarrow X\) の逆向きでしか成立しない
場合に \texttt{Invalid} を検討してください。
\(X\) が直接操作しにくい属性・指標であるという理由だけで
\texttt{Invalid} としてはいけません。

\textbf{【対象変数】}

\(X\):
\texttt{\{X\}}
（\texttt{\{X\_desc\}}）

\(Y\):
\texttt{\{Y\}}
（\texttt{\{Y\_desc\}}）

\(Z\):
\texttt{\{Z\}}

\textbf{【Zの役割】}

\texttt{<Appendix H の pattern-specific block>}

\textbf{【変数操作の定義】}

\texttt{<Appendix H の pattern-specific block>}

\textbf{【思考実験】}

\texttt{<Appendix H の pattern-specific block>}

\textbf{【判定ラベルの範囲】}

\texttt{<Appendix H の pattern-specific label set>}

\textbf{【判断基準】}

\texttt{<Appendix H の pattern-specific block>}

\textbf{【出力形式】}

以下のJSON形式で出力してください。
JSON以外の文字列は出力しないでください。

\begin{verbatim}
{
  "label": "<allowed judgment label>",
  "reason": "<judgment reason>"
}
\end{verbatim}

\end{quote}

\paragraph{Prompt G.2: English translation.}

\begin{quote}
\small

You are a \texttt{\{role\}} specializing in
\texttt{\{expertise\}}.
From the perspective of your specialty, judge the causal relation among
the given variables.

\textbf{[Task]}

Evaluate the causal relation \(X\rightarrow Y\) and determine whether
a direct effect from \(X\) to \(Y\) remains after considering the
third-variable hypothesis \(Z\).

A direct effect refers here to a causal pathway
\(X\rightarrow Y\) that does not operate through \(Z\).

Treat \(Z\) as an auxiliary hypothesis.
If the assumed role of \(Z\) clearly conflicts with known mechanisms,
general or domain knowledge, or temporal ordering, do not rely on that
hypothesis mechanically.

Do not use a mechanism supporting \(Y\rightarrow X\) as evidence for
\(X\rightarrow Y\).

First consider a plausible mechanism supporting
\(X\rightarrow Y\).
Consider \textsc{Invalid} when this mechanism is clearly inconsistent
with physical constraints, temporal ordering, or domain knowledge, or
when it can reasonably hold only in the reverse direction.
Do not select \textsc{Invalid} solely because \(X\) is difficult to
manipulate directly or is represented as an attribute or index.

\textbf{[Target variables]}

\(X\):
\texttt{\{X\}}
(\texttt{\{X\_desc\}})

\(Y\):
\texttt{\{Y\}}
(\texttt{\{Y\_desc\}})

\(Z\):
\texttt{\{Z\}}

\textbf{[Role of \(Z\)]}

\texttt{<pattern-specific block from Appendix H>}

\textbf{[Definition of conceptual intervention]}

\texttt{<pattern-specific block from Appendix H>}

\textbf{[Thought experiment]}

\texttt{<pattern-specific block from Appendix H>}

\textbf{[Available judgment labels]}

\texttt{<pattern-specific label set from Appendix H>}

\textbf{[Decision criteria]}

\texttt{<pattern-specific block from Appendix H>}

\textbf{[Output format]}

Output only a JSON object.

\begin{verbatim}
{
  "label": "<allowed judgment label>",
  "reason": "<judgment reason>"
}
\end{verbatim}

\end{quote}

\subsection{Final Selection Across Third Variables}
\label{app:phase2-final-selection}

When Phase~1 supplies multiple third-variable candidates, Phase~2
produces multiple candidate-level judgments for the same ordered pair.
The method therefore performs a second comparison in which each
persona is shown the candidate identifier, candidate-level label, and
a summarized reason for each \(Z\).

The task at this stage is not to generate another structural
hypothesis.
Instead, the persona selects the candidate-level judgment that best
explains whether a direct \(X\rightarrow Y\) pathway remains.
The selection emphasizes consistency with the
\(X\rightarrow Y\)-specific mechanism, the counterfactual intervention,
temporal ordering, and the structural role of \(Z\).

\paragraph{Prompt G.3: Original Japanese cross-candidate selection
template.}

\begin{quote}
\small

あなたは
\texttt{\{expertise\}}
を専門とする
\texttt{\{role\}}
です。

\textbf{【タスク】}

\(X\rightarrow Y\) への直接効果が成り立つのかを判定するため、
提示された \(Z\) 候補の中から、
\(X\rightarrow Y\) 固有の機序と直接経路の有無を
最も適切に説明しているものを1つ選んでください。

単に保守的・簡潔という理由ではなく、思考実験による
\(X\) から \(Y\) への影響を最も適切に評価している候補を
優先してください。
\(Y\rightarrow X\) のメカニズムを根拠にしている候補は
選択しないでください。

\textbf{【選択基準】}

\begin{itemize}
    \item
    \(X\rightarrow Y\) 固有の因果メカニズムで説明されているか。
    \item
    \(Z\) を考慮した後にも残る
    \(X\rightarrow Y\) の直接経路が理論的に支持されているか。
    \item
    \(Z\) が補助仮説であることを踏まえ、
    \(Z\) だけでは説明できない経路を検討しているか。
    \item
    candidate-level judgmentの理由が論理的に一貫しているか。
\end{itemize}

\textbf{【対象変数】}

\(X\):
\texttt{\{X\}}
（\texttt{\{X\_desc\}}）

\(Y\):
\texttt{\{Y\}}
（\texttt{\{Y\_desc\}}）

\textbf{【Z候補】}

\texttt{<candidate ID, candidate name, label, structural diagram,
and summarized reason for each candidate>}

\textbf{【出力形式】}

\begin{quote}
\begin{lstlisting}

{
  "selected_z": "<candidate ID>",
  "label": "<label attached to the selected candidate>",
  "reason": "<selection reason>"
}
\end{lstlisting}
\end{quote}

JSON以外の文字列は出力しないでください。
\end{quote}

\paragraph{Prompt G.4: English translation.}

\begin{quote}
\small

You are a \texttt{\{role\}} specializing in
\texttt{\{expertise\}}.

Select one of the presented third-variable candidates that most
appropriately explains whether a direct effect
\(X\rightarrow Y\) remains.

Do not prefer a candidate merely because it provides a simpler or more
conservative explanation.
Prefer the judgment that most appropriately evaluates the effect of a
counterfactual change in \(X\) on \(Y\).
Do not select a candidate whose justification relies on a mechanism
supporting \(Y\rightarrow X\).

Evaluate the candidates according to whether the explanation identifies
a mechanism specific to \(X\rightarrow Y\), whether a direct path
remains after accounting for \(Z\), whether alternative pathways not
explained by \(Z\) have been considered, and whether the
candidate-level reasoning is internally consistent.

Output only:

\begin{quote}
\begin{lstlisting}
{
  "selected_z": "<candidate ID>",
  "label": "<label attached to the selected candidate>",
  "reason": "<selection reason>"
}
\end{lstlisting}
\end{quote}

\end{quote}

The selected label becomes the Phase~2 output
\(\ell_{XY}\).
Only labels containing \textsc{Direct} are subsequently mapped to the
presence of the directed edge \(X\rightarrow Y\); all other labels are
mapped to its absence.

\section{Pattern-Specific Differences in Phase 2}
\label{app:phase2-differences}

Appendix~\ref{app:phase2-prompt} presents the common structure of the
Phase~2 direct-causality judgment.
This appendix specifies the condition-dependent blocks inserted into
that structure.
The experiments used the corresponding Japanese prompt text; the
descriptions below are English translations for readability.

Three input conditions are distinguished.
In the full condition, both the third-variable candidate \(Z\) and its
structural pattern are supplied by Phase~1.
In the no-predefined-candidate condition, neither \(Z\) nor its pattern
is specified in advance.
In the candidate-specified, pattern-unspecified condition, \(Z\) is
provided but its structural pattern is not.
The latter condition is used only in the corresponding comparison
experiment.

Across these conditions, a label containing \textsc{Direct} indicates
that a direct \(X\rightarrow Y\) pathway is judged to remain after the
relevant structural alternative has been considered.
A label without \textsc{Direct} indicates that the relation can be
explained without retaining that direct pathway.
\textsc{Invalid} is reserved for cases in which
\(X\rightarrow Y\) is strongly contradicted by temporal ordering,
physical or domain constraints, or is reasonably explainable only in
the reverse direction.

\subsection{Common Intervention and Directionality Rules}
\label{app:phase2-common-rules}

The implemented prompt defines manipulation of \(X\) conceptually
rather than restricting it to physically executable interventions.
Depending on the variable, the hypothetical intervention may correspond
to a design change, condition change, component replacement, policy or
environmental change, relocation, or replacement of the comparison
object.
Thus, an attribute, index, or measured quantity is not rejected solely
because its numerical value cannot be directly manipulated.

Before selecting a label, the persona is asked to identify a plausible
mechanism specific to \(X\rightarrow Y\) and check its consistency with
temporal ordering, physical or engineering principles, and domain
knowledge.
A mechanism supporting \(Y\rightarrow X\) must not be reused as
evidence for \(X\rightarrow Y\).

The supplied \(Z\) remains an auxiliary hypothesis.
If its assigned role clearly conflicts with domain knowledge or temporal
ordering, the persona may evaluate \(X\rightarrow Y\) without relying
on that hypothesis.
However, retaining a \textsc{Direct} label requires an independent
mechanism supporting \(X\rightarrow Y\), rather than merely rejecting
the supplied \(Z\).

\subsection{Pattern-Specific Blocks in the Full Condition}
\label{app:phase2-pattern-blocks}

When Phase~1 supplies both \(Z\) and its structural pattern, the
thought experiment is specialized as follows.

\paragraph{Chain.}
\(Z\) is treated as a mediator candidate,
\[
    X\rightarrow Z\rightarrow Y.
\]
The persona first specifies a conceptual intervention on \(X\), then
considers a counterfactual situation in which \(Z\) is externally held
fixed so that the mediated pathway through \(Z\) is blocked.
If changing \(X\) could still change \(Y\), the persona must further
consider whether that change can be explained by another indirect
pathway.
\textsc{Chain+Direct} is selected only when a mechanism specific to
\(X\rightarrow Y\), not explained by the mediated or another indirect
pathway, can be identified.
Otherwise, when the effect disappears after fixing \(Z\),
\textsc{Chain} is selected.
The available labels are
\[
    \{\textsc{Chain},
      \textsc{Chain+Direct},
      \textsc{Invalid}\}.
\]

\paragraph{Fork.}
\(Z\) is treated as a common-cause candidate,
\[
    X\leftarrow Z\rightarrow Y.
\]
The persona considers a counterfactual situation in which \(Z\) is
externally held fixed, thereby blocking variation attributable to the
hypothesized confounding path.
If changing \(X\) could still change \(Y\), the persona examines whether
the remaining change can instead be explained by another indirect
pathway.
\textsc{Fork+Direct} is selected only when a mechanism specific to
\(X\rightarrow Y\) remains after this assessment.
If fixing \(Z\) makes an effect of \(X\) on \(Y\) implausible,
\textsc{Fork} is selected.
The available labels are
\[
    \{\textsc{Fork},
      \textsc{Fork+Direct},
      \textsc{Invalid}\}.
\]

\paragraph{Collider.}
\(Z\) is treated as a collider candidate,
\[
    X\rightarrow Z\leftarrow Y.
\]
Unlike the Chain and Fork cases, the persona is instructed not to
condition on \(Z\), because stratifying, selecting, or fixing a
collider may induce a spurious association between \(X\) and \(Y\).
The thought experiment therefore asks whether a conceptual intervention
on \(X\) could change \(Y\) without conditioning on \(Z\).
\textsc{Collider+Direct} is selected only when such a change can be
supported by a mechanism specific to \(X\rightarrow Y\) that is not
explained by another indirect pathway.
If an apparent \(X\)--\(Y\) relation arises only after conditioning on
\(Z\), it is not treated as evidence for a direct effect.
The available labels are
\[
    \{\textsc{Collider},
      \textsc{Collider+Direct},
      \textsc{Invalid}\}.
\]

Table~\ref{tab:phase2-pattern-summary} summarizes the operative
differences.

\begin{table*}[t]
\centering
\small
\begin{tabular}{llll}
\hline
Pattern & Structural hypothesis & Treatment of \(Z\) & Labels \\
\hline
Chain
& \(X\rightarrow Z\rightarrow Y\)
& Hold \(Z\) fixed
& Chain / Chain+Direct / Invalid \\

Fork
& \(X\leftarrow Z\rightarrow Y\)
& Hold \(Z\) fixed
& Fork / Fork+Direct / Invalid \\

Collider
& \(X\rightarrow Z\leftarrow Y\)
& Do not condition on \(Z\)
& Collider / Collider+Direct / Invalid \\
\hline
\end{tabular}
\caption{Pattern-specific conditions used in the Phase~2 thought
experiment.}
\label{tab:phase2-pattern-summary}
\end{table*}

\subsection{No-Predefined-Candidate Condition}
\label{app:phase2-no-candidate}

When no third-variable candidate is supplied, the persona must first
hypothesize one plausible \(Z\) relevant to the ordered pair
\((X,Y)\).
It then assigns that candidate one of the three structural roles:
\[
    \textsc{Fork},
    \textsc{Chain},
    \textsc{Collider}.
\]

The persona then performs the corresponding thought experiment:
\(Z\) is held fixed for a Fork or Chain hypothesis and is not
conditioned on for a Collider hypothesis.
As in the full condition, the persona must distinguish a mechanism
specific to \(X\rightarrow Y\) from an effect that depends on changing
\(Z\) or can be explained through an indirect pathway.

Because neither \(Z\) nor its structural pattern is given in advance,
all seven labels are available:
\begin{align*}
  \{&\textsc{Fork},
  \textsc{Fork+Direct},
  \textsc{Chain},\\
  &\textsc{Chain+Direct},
  \textsc{Collider},\\
  &\textsc{Collider+Direct},
  \textsc{Invalid}\}.
  \end{align*}

The implemented output additionally records the third variable and
structural pattern hypothesized by the persona:

\begin{quote}
\begin{lstlisting}
{
  "assumed_z": "<hypothesized third variable>",
  "assumed_pattern": "Fork|Chain|Collider",
  "label": "<allowed judgment label>",
  "reason": "<judgment reason>"
}
\end{lstlisting}
\end{quote}

For a label containing \textsc{Direct}, the prompt requires the persona
to identify an \(X\rightarrow Y\)-specific mechanism that remains after
the hypothesized structural alternative has been considered and that is
consistent with temporal and domain constraints.

\subsection{Candidate-Specified, Pattern-Unspecified Condition}
\label{app:phase2-pattern-unspecified}

In this comparison condition, the identity of \(Z\) is supplied, but
its structural role is withheld.
The persona therefore does not generate a new third-variable candidate.
Instead, it first determines whether the supplied \(Z\) is most
plausibly interpreted as
 \begin{gather*}
  X\leftarrow Z\rightarrow Y,
  \qquad
  X\rightarrow Z\rightarrow Y,\\
  \text{or}\qquad
  X\rightarrow Z\leftarrow Y.
  \end{gather*}

After selecting the structural hypothesis, the persona performs the
corresponding thought experiment.
For a Fork hypothesis, \(Z\) is held fixed to block the hypothesized
confounding path.
For a Chain hypothesis, \(Z\) is held fixed to block the hypothesized
mediated path.
For a Collider hypothesis, the judgment is made without conditioning
on \(Z\).

All seven judgment labels remain available:
 \{\textsc{Fork}, \textsc{Fork+Direct}, \textsc{Chain}, \textsc{Chain+Direct},
  \textsc{Collider}, \textsc{Collider+Direct}, \textsc{Invalid}\}.

The output contains the selected label and its justification:

\begin{quote}
\begin{lstlisting}
    
{
  "label": "<allowed judgment label>",
  "reason": "<selected structural hypothesis and
             X->Y-specific justification>"
}
\end{lstlisting}
\end{quote}

The prompt requires the reason to state which structural hypothesis was
adopted for the supplied \(Z\), even though that hypothesis is not
returned as a separate JSON field.

\subsection{Use in Subsequent Phase 2 Judgments}
\label{app:phase2-difference-reuse}

The blocks above define the structural semantics used in the initial
candidate-level judgment.
The same pattern interpretation and label-specific decision criteria
are reused in the subsequent review and final-decision prompts, with
shortened descriptions where appropriate.
Thus, the later exchanges may revise a judgment, but they do not change
the meaning of Chain, Fork, Collider, or their corresponding
\textsc{Direct} labels.

\section{Phase 3 Bidirectional-Edge Resolution and Cycle Removal}
\label{app:phase3}

Phase~3 reconciles the directed-edge candidates obtained from
independent ordered-pair judgments in Phase~2.
Because \((X,Y)\) and \((Y,X)\) are evaluated separately, the
intermediate graph may contain both opposing edges for the same
unordered pair and may also contain directed cycles.

Phase~3 therefore applies two graph-level corrections.
First, bidirectional conflicts are resolved through a constrained
LLM-based binary comparison.
Second, if directed cycles remain, candidate acyclic graphs are
compared using a BIC-based procedure with the observational data.
The latter step is mechanical and does not involve further LLM
judgment.

If cycle removal completes and produces an acyclic graph, that graph is
adopted.
Otherwise, the graph obtained after bidirectional-edge resolution is
retained as the adopted estimate.
Accordingly, the fallback output is not necessarily a DAG.

\subsection{Overall Phase 3 Workflow}
\label{app:phase3-workflow}

Let \(G\) denote the intermediate directed graph obtained by mapping the
Phase~2 labels to edges.
Labels containing \textsc{Direct} are mapped to the presence of
\(X\rightarrow Y\); labels without \textsc{Direct} and
\textsc{Invalid} are mapped to its absence.

\begin{algorithm}[t]
    \caption{Structural reconciliation in Phase~3}
    \label{alg:phase3}
    \small
    \begin{algorithmic}[1]
        \Require Intermediate directed graph $G$; variable descriptions $d$;
stored-reason map $r$ for the surviving directed edges;
observational data $D$; persona set $P=\{p_1,p_2,p_3\}$

        \Ensure
        Adopted directed graph \(G^{*}\)

        \State \(G'\gets G\)

        \If{\(G'\) contains a bidirectional pair}
            \State
            \(G'\gets
            \Call{ResolveBidirectionalEdges}
            {G',\mathbf{d},r,\mathcal{P}}\)
        \EndIf

        \State \(G_{\mathrm{bi}}\gets G'\)

        \If{\(G_{\mathrm{bi}}\) contains a directed cycle}
            \State Attempt BIC-based cycle removal using \(D\)

            \If{an acyclic output is successfully produced}
                \State Set \(G'\) to the selected acyclic graph
            \Else
                \State \(G'\gets G_{\mathrm{bi}}\)
            \EndIf
        \Else
            \State \(G'\gets G_{\mathrm{bi}}\)
        \EndIf

        \State \(G^{*}\gets G'\)
        \State \Return \(G^{*}\)
    \end{algorithmic}
\end{algorithm}

\subsection{Bidirectional-Edge Resolution}
\label{app:phase3-bidirectional}

Suppose both
\[
    X\rightarrow Y
    \qquad\text{and}\qquad
    Y\rightarrow X
\]
survive the Phase~2 mapping.
For this unordered pair, each of the three personas is asked to compare
the two surviving directional hypotheses.

Each persona receives the natural-language descriptions of \(X\) and
\(Y\), the stored reason supporting \(X\rightarrow Y\), and the stored
reason supporting \(Y\rightarrow X\).
The Phase~1 third-variable candidates are not regenerated or presented
again, and the Phase~2 final labels themselves are not included in the
current direction-selection prompt.

The task is explicitly binary.
Each persona must select exactly one of
\(X\rightarrow Y\) and \(Y\rightarrow X\).
The comparison is based on temporal ordering, known causal mechanisms,
general knowledge, and domain knowledge.
A mechanism supporting the reverse direction may not be reused as
evidence for the selected direction.

With three valid binary responses, one direction necessarily receives
at least two votes and the opposite direction is removed.
A response that cannot be parsed as one of the two permitted
directions does not contribute a directional vote; if neither
direction receives two votes, the bidirectional pair is left
unchanged by this step.

\begin{algorithm}[t]
    \caption{Bidirectional-edge resolution}
    \label{alg:bidirectional-resolution}
    \small
    \begin{algorithmic}[1]
        \Require
        Directed graph \(G\);
        variable descriptions \(\mathbf{d}\);
        stored reasons \(r_{XY}\) and \(r_{YX}\);
        persona set \(\mathcal{P}=\{p_1,p_2,p_3\}\)

        \Ensure
        Updated graph \(G'\)

        \State \(G'\gets G\)

        \ForAll{unordered pairs \(\{X,Y\}\) such that
        \(X\rightarrow Y\in G'\) and
        \(Y\rightarrow X\in G'\)}

            \ForAll{\(p\in\mathcal{P}\)}
                \State Present the descriptions of \(X\) and \(Y\)
                \State Present \(r_{XY}\) and \(r_{YX}\)
                \State Ask \(p\) to select exactly one direction
                \State Record the selected direction \(v_p\)
            \EndFor

            \State
            \(c_{XY}\gets
            \sum_{p\in\mathcal{P}}
            \mathbf{1}[v_p=X\rightarrow Y]\)

            \State
            \(c_{YX}\gets
            \sum_{p\in\mathcal{P}}
            \mathbf{1}[v_p=Y\rightarrow X]\)

            \If{$c_{XY} \geq 2$}
                \State Remove $Y \rightarrow X$ from $G'$
            \ElsIf{$c_{YX} \geq 2$}
                \State Remove $X \rightarrow Y$ from $G'$
            \Else
                \State Leave both directions unchanged
            \EndIf
        \EndFor

        \State \Return \(G'\)
    \end{algorithmic}
\end{algorithm}

\subsubsection{Bidirectional-Edge Resolution Prompt}

The experiments used the Japanese prompt below.
The English version is provided only as a translation.

\paragraph{Prompt I.1: Original Japanese prompt.}

\textbf{Persona system message}

\begin{quote}
\small
あなたは
\texttt{\{expertise\}}
を専門とする
\texttt{\{role\}}
です。
因果の方向を二択で判定してください。
\end{quote}

\textbf{Direction-selection system message}

\begin{quote}
\small
変数 \(X\) と \(Y\) の間で、以下の候補方向のうちどちらが
より妥当かを評価してください。

\begin{itemize}
    \item \(X\rightarrow Y\)
    \item \(Y\rightarrow X\)
\end{itemize}

判定の際は以下を考慮してください。

\begin{itemize}
    \item 時間順序との整合性
    \item 既知の因果メカニズム（法則や関係式など）
    \item 一般知識・ドメイン知識との整合性
\end{itemize}

\(Y\rightarrow X\) のメカニズムを
\(X\rightarrow Y\) の根拠として流用してはいけません。

両方向それぞれのメカニズムを想定した上で、
より妥当と考えられる方向を必ずどちらか一方選択してください。
「判断できない」「どちらでもない」という回答は認められません。

出力はJSON形式のみとし、それ以外の文字列は含めないで
ください。
\end{quote}

\textbf{User message}

\begin{quote}
\small
【対象変数】

\(X\):
\texttt{\{X\}}:
\texttt{\{X\_desc\}}

\(Y\):
\texttt{\{Y\}}:
\texttt{\{Y\_desc\}}

【候補1】
\texttt{\{X\}}
\(\rightarrow\)
\texttt{\{Y\}}

根拠:
\texttt{\{forward\_reason\}}

【候補2】
\texttt{\{Y\}}
\(\rightarrow\)
\texttt{\{X\}}

根拠:
\texttt{\{backward\_reason\}}

上記の情報に基づき、より妥当な方向を必ず1つ選択し、
以下のJSON形式で出力してください。

\begin{lstlisting}
{
  "direction": "X->Y" または "Y->X",
  "reason": "選択理由（日本語30--120字）。
             反事実操作と時間順序との整合を具体的に述べること。
             選択した方向に固有のメカニズム
             （定性的な機構の説明でも可）を1つ挙げること"
}
\end{lstlisting}
\end{quote}

\paragraph{Prompt I.2: English translation.}

\textbf{Persona system message}

\begin{quote}
\small
You are a \texttt{\{role\}} specializing in
\texttt{\{expertise\}}.
Judge the causal direction as a binary choice.
\end{quote}

\textbf{Direction-selection system message}

\begin{quote}
\small
Evaluate which of the following candidate directions between
\(X\) and \(Y\) is more plausible:

\begin{itemize}
    \item \(X\rightarrow Y\)
    \item \(Y\rightarrow X\)
\end{itemize}

Consider temporal ordering, known causal mechanisms including laws or
functional relations, and general and domain knowledge.

Do not reuse a mechanism supporting \(Y\rightarrow X\) as evidence for
\(X\rightarrow Y\).

After considering the mechanisms for both directions, select exactly
one of the two directions.
Responses such as ``cannot determine'' or ``neither direction'' are
not permitted.

Output only JSON.
\end{quote}

\textbf{User message}

\begin{quote}
\small
[Target variables]

\(X\):
\texttt{\{X\}}:
\texttt{\{X\_desc\}}

\(Y\):
\texttt{\{Y\}}:
\texttt{\{Y\_desc\}}

[Candidate 1]

\texttt{\{X\}}
\(\rightarrow\)
\texttt{\{Y\}}

Reason:
\texttt{\{forward\_reason\}}

[Candidate 2]

\texttt{\{Y\}}
\(\rightarrow\)
\texttt{\{X\}}

Reason:
\texttt{\{backward\_reason\}}

Select exactly one direction and output:

\begin{lstlisting}
{
  "direction": "X->Y" or "Y->X",
  "reason": "State the reason in 30--120 Japanese characters.
             Describe consistency with a counterfactual intervention
             and temporal ordering, and identify one mechanism
             specific to the selected direction."
}
\end{lstlisting}
\end{quote}

\subsection{Cycle Removal}
\label{app:phase3-cycle}

Bidirectional-edge resolution removes the opposite edge whenever one
direction receives at least two valid votes.
If neither direction reaches this threshold, the bidirectional pair is
left unchanged.
Consequently, the graph entering cycle removal may contain residual
bidirectional pairs as well as other directed cycles.
If one or more directed cycles remain, the method attempts a post-hoc
BIC-based graph-selection procedure using the observational data.

The procedure constructs candidate graphs by removing edges from the
cyclic graph until acyclic candidates are obtained.
Only acyclic candidates are scored.
Among the candidate DAGs produced by this search, the graph with the
lowest BIC is selected.

This step does not introduce new edges and does not use LLM judgments.
It only selects among subgraphs of the graph obtained after
bidirectional-edge resolution.

\subsubsection{Cycle-Removal Algorithm}

\begin{algorithm}[t]
    \caption{Post-hoc cycle removal by BIC minimization}
    \label{alg:cycle-removal}
    \small
    \begin{algorithmic}[1]
        \Require
        Graph \(G_{\mathrm{bi}}\) after bidirectional-edge resolution;
        observational data \(D\)

        \Ensure
        Selected acyclic graph \(G^{\dagger}\), if the procedure
        completes

        \State Detect directed cycles in \(G_{\mathrm{bi}}\)

        \State Enumerate candidate graphs obtained by removing edge
        combinations sufficient to break the detected cycles

        \State Discard candidate graphs that remain cyclic

        \State Let \(\mathcal{G}_{\mathrm{DAG}}\) denote the resulting
        set of acyclic candidates

        \ForAll{\(H\in\mathcal{G}_{\mathrm{DAG}}\)}
            \State Compute \(\operatorname{BIC}(H;D)\)
        \EndFor

        \State
        \(G^{\dagger}
        \gets
        \arg\min_{H\in\mathcal{G}_{\mathrm{DAG}}}
        \operatorname{BIC}(H;D)\)

        \State \Return \(G^{\dagger}\)
    \end{algorithmic}
\end{algorithm}

\subsubsection{BIC Computation}

For a candidate graph \(H\), the scoring rule used in this study is
\[
    \operatorname{BIC}(H;D)
    =
    N\log\left(\frac{\operatorname{RSS}(H;D)}{N}\right)
    +
    k_H\log N,
\]
where \(N\) is the number of observations,
\(\operatorname{RSS}(H;D)\) is the total residual sum of squares across
the variables under the candidate structure, and \(k_H\) is the number
of directed edges in \(H\) under the scoring convention used here.
Lower values are preferred.

The same standardized observational data used for the statistical
baselines are used in this reconciliation step.
The BIC criterion is used as a common mechanical ranking rule among
candidate acyclic subgraphs; it should not be interpreted as an
assumption that every dataset exactly follows a linear-Gaussian
data-generating process.

\subsection{Role of Phase 3}
\label{app:phase3-role}

Phase~3 is a structural reconciliation stage rather than an additional
knowledge-generation stage.
Bidirectional-edge resolution compares two already surviving
directional hypotheses through a constrained binary vote.
Cycle removal subsequently uses observational data only when the
resulting graph remains cyclic.

Thus, the two corrections serve different purposes:
the first resolves directional conflicts among local LLM judgments,
whereas the second attempts to satisfy the global DAG constraint by
selecting among edge-deleted subgraphs.
When the latter procedure does not produce an acyclic output, the graph
after bidirectional-edge resolution is retained as the adopted
estimate.

\section{Formal Definitions of Evaluation Metrics}
\label{app:metrics}

This appendix provides the formal definitions of the evaluation metrics
used in this study.
We evaluate each estimated graph by comparing its directed edge set
with the directed edge set of the adopted reference graph.

Let \(V\) be the set of variables and let \(|V|=n\).
Let \(E^\ast \subseteq V \times V\) denote the directed edge set of the
reference graph, and let \(\hat{E} \subseteq V \times V\) denote the
directed edge set estimated by a method.
Self-loops are not allowed in either graph.

For edge-level classification metrics, we define the universe of
possible directed non-self-loop edges as
\[
U = \{(X,Y)\in V\times V : X\neq Y\}.
\]

The confusion-count terms are defined as
\[
\begin{aligned}
TP &= |\hat{E}\cap E^\ast|,\\
FP &= |\hat{E}\setminus E^\ast|,\\
FN &= |E^\ast\setminus \hat{E}|,\\
TN &= |U\setminus(\hat{E}\cup E^\ast)|.
\end{aligned}
\]

A predicted edge \(X\rightarrow Y\) is counted as a true positive only
when the same directed edge is present in the reference graph.
If the reference graph contains \(X\rightarrow Y\) but the estimated
graph instead contains \(Y\rightarrow X\), the mismatch contributes
one false negative for \(X\rightarrow Y\) and one false positive for
\(Y\rightarrow X\).

Precision is defined as
\[
\mathrm{Precision}
=
\frac{TP}{TP+FP}.
\]

Recall is defined as
\[
\mathrm{Recall}
=
\frac{TP}{TP+FN}.
\]

F1 is the harmonic mean of Precision and Recall:
\[
\mathrm{F1}
=
\frac{
2\,\mathrm{Precision}\,\mathrm{Recall}
}{
\mathrm{Precision}+\mathrm{Recall}
}.
\]

The false-positive rate is defined as
\[
\mathrm{FPR}
=
\frac{FP}{FP+TN},
\]
and the false-negative rate as
\[
\mathrm{FNR}
=
\frac{FN}{FN+TP}.
\]

Because
\[
\mathrm{Recall}
=
\frac{TP}{TP+FN},
\]
FNR and Recall satisfy
\[
\mathrm{FNR}=1-\mathrm{Recall},
\]
provided that \(TP+FN>0\).

Normalized Hamming distance is computed at the adjacency-matrix level.
Let \(A^\ast\) and \(\hat{A}\) denote the \(n\times n\) adjacency
matrices of the reference and estimated graphs, respectively.
Then
\[
\mathrm{NHD}
=
\frac{1}{n^2}
\sum_{i=1}^{n}
\sum_{j=1}^{n}
\mathbf{1}
\left\{
\hat{A}_{ij}\neq A^\ast_{ij}
\right\}.
\]

Because self-loops are excluded from both graphs, the diagonal entries
of both adjacency matrices are zero and do not contribute to the
numerator.
Equivalently,
\[
\mathrm{NHD}
=
\frac{FP+FN}{n^2}.
\]

Thus, the edge-level classification metrics are defined over the
non-self-loop directed-edge universe \(U\), whereas NHD is the
adjacency-matrix Hamming distance normalized by \(n^2\).

When a denominator is zero, the corresponding metric is undefined in
principle.
In the reported experiments, each adopted reference graph contains at
least one directed edge.
If a method outputs no predicted edges, Precision is set to zero for
reporting purposes, and F1 is set to zero when
\(\mathrm{Precision}+\mathrm{Recall}=0\).

For methods or conditions evaluated over multiple runs, each metric is
computed separately for each run and then summarized by the mean and
sample standard deviation.

\section{Stage-wise Retention of Reference Edges}
\label{app:edge-diagnostics}

This appendix reports stage-wise reference-edge retention for the
GPT-5.4 \textsc{Full} and \textsc{No Phase 1} conditions.
Unlike a selected-run illustration, the analysis uses the same three
independent runs per dataset as the main experiments and summarizes
the results by the mean and sample standard deviation.

For each run, we first measure the proportion of reference edges that
survive the Phase~2 direct-edge judgment.
A reference edge \(X\rightarrow Y\) is counted as retained after
Phase~2 when its final pair-level label contains
\texttt{+Direct}.
We denote this quantity by

\[
R_{\mathrm{P2}}
=
\frac{|E^\ast_{\mathrm{P2}}|}{|E^\ast|},
\]
where \(E^\ast_{\mathrm{P2}}\) denotes the set of reference edges
retained after Phase~2.

Final retention is the Recall of the adopted graph,

\[
R_{\mathrm{final}}
=
\frac{TP}{TP+FN}.
\]

We define the additional reference-edge loss introduced during
Phase~3 structural reconciliation as

\[
L_{\mathrm{P3}}
=
R_{\mathrm{P2}} - R_{\mathrm{final}}.
\]

The loss term includes any reference edge that survives the Phase~2
mapping but is absent from the adopted graph after bidirectional-edge
resolution and, when applicable, cycle reconciliation.
It therefore localizes the loss to Phase~3 as a whole rather than
attributing it to a particular reconciliation substep.

\begin{table*}[t]
\centering
\small
\setlength{\tabcolsep}{7pt}
\begin{tabular}{llccc}
\toprule
Dataset &
Condition &
Phase~2 retention &
Final Recall &
Phase~3 loss \\
\midrule
Auto-MPG
& Full
& \(0.800 \pm 0.000\)
& \(0.800 \pm 0.000\)
& \(0.000 \pm 0.000\) \\
&
No Phase~1
& \(0.400 \pm 0.000\)
& \(0.400 \pm 0.000\)
& \(0.000 \pm 0.000\) \\
\midrule
DWD
& Full
& \(0.944 \pm 0.096\)
& \(0.944 \pm 0.096\)
& \(0.000 \pm 0.000\) \\
&
No Phase~1
& \(0.000 \pm 0.000\)
& \(0.000 \pm 0.000\)
& \(0.000 \pm 0.000\) \\
\midrule
Sachs
& Full
& \(0.702 \pm 0.080\)
& \(0.526 \pm 0.053\)
& \(0.175 \pm 0.061\) \\
&
No Phase~1
& \(0.070 \pm 0.030\)
& \(0.070 \pm 0.030\)
& \(0.000 \pm 0.000\) \\
\bottomrule
\end{tabular}
\caption{
Stage-wise retention of reference edges under GPT-5.4.
Phase~2 retention is the proportion of reference edges whose
pair-level Phase~2 label contains \texttt{+Direct}.
Final Recall is measured on the adopted graph.
Phase~3 loss is computed separately for each run as the difference
between Phase~2 retention and final Recall, and the table reports
the mean and sample standard deviation over three runs.
}
\label{tab:stagewise-retention}
\end{table*}

Under \textsc{No Phase 1}, Phase~2 retention and final Recall are
identical on all three datasets.
Thus, no additional reference-edge loss is introduced during
Phase~3 in these runs; the final omissions are already determined by
the direct-edge judgments made in Phase~2.

The \textsc{Full} condition shows the same pattern on Auto-MPG and
DWD, where Phase~3 introduces no additional reference-edge loss.
Sachs differs substantially.
The Full condition retains \(70.2\%\) of the reference edges after
Phase~2 but \(52.6\%\) in the adopted graph, corresponding to an
additional mean loss of \(17.5\) percentage points during
Phase~3 reconciliation.

These results localize the main Full--No Phase~1 difference primarily
to the direct-edge judgment stage.
At the same time, the Sachs Full result shows that graph-level
reconciliation can introduce a non-negligible additional source of
omission when the intermediate graph requires greater structural
reconciliation.

\section{Execution Cost Details}
\label{app:execution-cost}

MaSCoD uses staged processing and multiple persona-based judgments,
resulting in substantially more LLM calls than a one-shot pairwise
procedure.
Table~\ref{tab:execution-cost} summarizes the execution cost of the
primary GPT-5.4 \textsc{Full} experiments.
For each dataset, we report the total number of logged LLM calls,
processed tokens, and cumulative LLM-call time, summarized by the mean
and sample standard deviation over the same three runs used in the
main evaluation.

The time measure is the sum of recorded LLM-call durations rather than
end-to-end wall-clock runtime.
Token counts are obtained from the usage information recorded for the
LLM calls.

\begin{table*}[t]
\centering
\small
\setlength{\tabcolsep}{6pt}
\begin{tabular}{lrrrr}
\toprule
Dataset &
Pairs &
LLM calls &
Tokens (M) &
Cumulative LLM time (s) \\
\midrule
Auto-MPG
& 20
& \(1264 \pm 47\)
& \(3.367 \pm 0.112\)
& \(10736 \pm 789\) \\
DWD
& 30
& \(1863 \pm 50\)
& \(5.175 \pm 0.146\)
& \(18883 \pm 1815\) \\
Sachs
& 110
& \(8373 \pm 43\)
& \(26.369 \pm 0.019\)
& \(88166 \pm 2169\) \\
\bottomrule
\end{tabular}
\caption{
Execution cost of the GPT-5.4 \textsc{Full} condition.
LLM calls, processed tokens, and cumulative LLM-call time are reported
as mean \(\pm\) sample standard deviation over three runs.
Pairs denotes ordered non-self variable pairs.
}
\label{tab:execution-cost}
\end{table*}

The measured cost increases substantially with the size of the
evaluated variable set.
From Auto-MPG to Sachs, the number of ordered pairs increases from
20 to 110, while the mean number of LLM calls increases from
approximately 1.26k to 8.37k and mean token usage from approximately
3.37M to 26.37M.
These measurements should be interpreted descriptively rather than as
an empirical complexity analysis, because call and token usage also
depend on the number of third-variable candidates retained for each
pair and on the iterative judgment steps invoked during processing.

MaSCoD is therefore considerably more inference-intensive than
one-shot pairwise prompting.
Its current implementation targets the small-to-medium exploratory
candidate-graph settings evaluated in this study rather than
computational efficiency as a primary objective.

\paragraph{Compute reporting.}
GPT-5.4 and GPT-4o were accessed as hosted models through Azure
OpenAI Service. Their parameter counts and server-side hardware
configurations are not disclosed to us, so GPU-hour estimates are not
available. We therefore report observable API-level quantities,
including model identifiers, generation settings, request counts,
token usage, and cumulative LLM-call time where recorded.

\section{Complete Evaluation Results}
\label{app:full-results}

This appendix reports the complete evaluation metrics underlying the
main-text results.
For LLM-based methods and ablation conditions, we report
the mean and sample standard deviation over three evaluations.
The repetition protocol differs for the partial-information
conditions: all three Identity-only evaluations and all
three Pattern-only evaluations reuse masked versions of
the Phase~1 outputs from Full run~1.
Their dispersion estimates therefore characterize
downstream variability conditional on this fixed source
context and do not include variability across regenerated
Phase~1 contexts.
Statistical baselines were evaluated once and are therefore reported
without dispersion estimates.
NHD follows the adjacency-matrix definition given in
Section~\ref{sec:experimental-setup}.

\subsection{Complete Results for RQ1}

Tables~\ref{tab:full-rq1-auto}--\ref{tab:full-rq1-sachs} report
Precision, Recall, F1, FPR, FNR, and NHD for the methods compared in
RQ1.

\begin{table*}[t]
\centering
\scriptsize
\setlength{\tabcolsep}{3.5pt}
\begin{tabular}{llcccccc}
\toprule
Backbone & Method &
Precision & Recall & F1 & FPR & FNR & NHD \\
\midrule
-- &
PC &
0.143 & 0.200 & 0.167 & 0.400 & 0.800 & 0.400 \\
-- &
DirectLiNGAM &
0.111 & 0.200 & 0.143 & 0.533 & 0.800 & 0.480 \\
-- &
Exact Search &
0.500 & 0.800 & 0.615 & 0.267 & 0.200 & 0.200 \\
\midrule
GPT-5.4 &
MaSCoD &
\(0.487 \pm 0.073\) &
\(0.800 \pm 0.000\) &
\(0.603 \pm 0.055\) &
\(0.289 \pm 0.077\) &
\(0.200 \pm 0.000\) &
\(0.213 \pm 0.046\) \\
GPT-5.4 &
LLM-KBCI &
\(0.500 \pm 0.000\) &
\(0.800 \pm 0.000\) &
\(0.615 \pm 0.000\) &
\(0.267 \pm 0.000\) &
\(0.200 \pm 0.000\) &
\(0.200 \pm 0.000\) \\
GPT-5.4 &
MAC &
\(0.603 \pm 0.055\) &
\(0.800 \pm 0.000\) &
\(0.687 \pm 0.035\) &
\(0.178 \pm 0.038\) &
\(0.200 \pm 0.000\) &
\(0.147 \pm 0.023\) \\
\midrule
GPT-4o &
MaSCoD &
\(0.444 \pm 0.000\) &
\(0.800 \pm 0.000\) &
\(0.571 \pm 0.000\) &
\(0.333 \pm 0.000\) &
\(0.200 \pm 0.000\) &
\(0.240 \pm 0.000\) \\
GPT-4o &
LLM-KBCI &
\(0.566 \pm 0.063\) &
\(0.867 \pm 0.116\) &
\(0.684 \pm 0.078\) &
\(0.222 \pm 0.038\) &
\(0.133 \pm 0.116\) &
\(0.160 \pm 0.040\) \\
GPT-4o &
MAC &
\(0.490 \pm 0.086\) &
\(0.800 \pm 0.000\) &
\(0.605 \pm 0.067\) &
\(0.289 \pm 0.102\) &
\(0.200 \pm 0.000\) &
\(0.213 \pm 0.061\) \\
\bottomrule
\end{tabular}
\caption{Complete RQ1 results on Auto-MPG. LLM-based values are
mean \(\pm\) sample standard deviation over three runs.}
\label{tab:full-rq1-auto}
\end{table*}

\begin{table*}[t]
\centering
\scriptsize
\setlength{\tabcolsep}{3.5pt}
\begin{tabular}{llcccccc}
\toprule
Backbone & Method &
Precision & Recall & F1 & FPR & FNR & NHD \\
\midrule
-- &
PC &
0.125 & 0.167 & 0.143 & 0.292 & 0.833 & 0.333 \\
-- &
DirectLiNGAM &
0.154 & 0.333 & 0.211 & 0.458 & 0.667 & 0.417 \\
-- &
Exact Search &
0.273 & 0.500 & 0.353 & 0.333 & 0.500 & 0.306 \\
\midrule
GPT-5.4 &
MaSCoD &
\(0.567 \pm 0.029\) &
\(0.944 \pm 0.096\) &
\(0.708 \pm 0.042\) &
\(0.181 \pm 0.024\) &
\(0.056 \pm 0.096\) &
\(0.130 \pm 0.016\) \\
GPT-5.4 &
LLM-KBCI &
\(0.387 \pm 0.077\) &
\(0.667 \pm 0.167\) &
\(0.488 \pm 0.099\) &
\(0.264 \pm 0.048\) &
\(0.333 \pm 0.167\) &
\(0.232 \pm 0.042\) \\
GPT-5.4 &
MAC &
\(0.622 \pm 0.038\) &
\(0.556 \pm 0.096\) &
\(0.586 \pm 0.070\) &
\(0.083 \pm 0.000\) &
\(0.444 \pm 0.096\) &
\(0.130 \pm 0.016\) \\
\midrule
GPT-4o &
MaSCoD &
\(0.524 \pm 0.041\) &
\(0.667 \pm 0.000\) &
\(0.586 \pm 0.025\) &
\(0.153 \pm 0.024\) &
\(0.333 \pm 0.000\) &
\(0.157 \pm 0.016\) \\
GPT-4o &
LLM-KBCI &
\(0.548 \pm 0.041\) &
\(0.667 \pm 0.000\) &
\(0.601 \pm 0.025\) &
\(0.139 \pm 0.024\) &
\(0.333 \pm 0.000\) &
\(0.148 \pm 0.016\) \\
GPT-4o &
MAC &
\(0.532 \pm 0.122\) &
\(0.556 \pm 0.096\) &
\(0.543 \pm 0.109\) &
\(0.125 \pm 0.042\) &
\(0.444 \pm 0.096\) &
\(0.157 \pm 0.042\) \\
\bottomrule
\end{tabular}
\caption{Complete RQ1 results on DWD. LLM-based values are
mean \(\pm\) sample standard deviation over three runs.}
\label{tab:full-rq1-dwd}
\end{table*}

\begin{table*}[t]
\centering
\scriptsize
\setlength{\tabcolsep}{3.5pt}
\begin{tabular}{llcccccc}
\toprule
Backbone & Method &
Precision & Recall & F1 & FPR & FNR & NHD \\
\midrule
-- &
PC &
0.385 & 0.526 & 0.444 & 0.176 & 0.474 & 0.207 \\
-- &
DirectLiNGAM &
0.278 & 0.526 & 0.364 & 0.286 & 0.474 & 0.289 \\
-- &
Exact Search &
0.303 & 0.526 & 0.385 & 0.253 & 0.474 & 0.265 \\
\midrule
GPT-5.4 &
MaSCoD &
\(0.348 \pm 0.039\) &
\(0.526 \pm 0.053\) &
\(0.417 \pm 0.032\) &
\(0.209 \pm 0.044\) &
\(0.474 \pm 0.053\) &
\(0.231 \pm 0.030\) \\
GPT-5.4 &
LLM-KBCI &
\(0.657 \pm 0.042\) &
\(0.526 \pm 0.053\) &
\(0.582 \pm 0.017\) &
\(0.059 \pm 0.017\) &
\(0.474 \pm 0.053\) &
\(0.118 \pm 0.005\) \\
GPT-5.4 &
MAC &
\(0.747 \pm 0.032\) &
\(0.316 \pm 0.053\) &
\(0.443 \pm 0.058\) &
\(0.022 \pm 0.000\) &
\(0.684 \pm 0.053\) &
\(0.124 \pm 0.008\) \\
\midrule
GPT-4o &
MaSCoD &
\(0.344 \pm 0.031\) &
\(0.684 \pm 0.105\) &
\(0.456 \pm 0.033\) &
\(0.275 \pm 0.058\) &
\(0.316 \pm 0.105\) &
\(0.256 \pm 0.033\) \\
GPT-4o &
LLM-KBCI &
\(0.590 \pm 0.041\) &
\(0.298 \pm 0.030\) &
\(0.395 \pm 0.022\) &
\(0.044 \pm 0.011\) &
\(0.702 \pm 0.030\) &
\(0.143 \pm 0.005\) \\
GPT-4o &
MAC &
\(0.645 \pm 0.081\) &
\(0.281 \pm 0.030\) &
\(0.390 \pm 0.036\) &
\(0.033 \pm 0.011\) &
\(0.719 \pm 0.030\) &
\(0.138 \pm 0.010\) \\
\bottomrule
\end{tabular}
\caption{Complete RQ1 results on Sachs. LLM-based values are
mean \(\pm\) sample standard deviation over three runs.}
\label{tab:full-rq1-sachs}
\end{table*}

\subsection{Complete Results for RQ2}

Tables~\ref{tab:full-ablation-gpt54} and
\ref{tab:full-ablation-gpt4o} report the complete Phase~1 ablation
results for GPT-5.4 and GPT-4o, respectively.

\begin{table*}[t]
\centering
\scriptsize
\setlength{\tabcolsep}{3.5pt}
\begin{tabular}{llcccccc}
\toprule
Dataset & Condition &
Precision & Recall & F1 & FPR & FNR & NHD \\
\midrule
Auto-MPG &
Full &
\(0.487 \pm 0.073\) &
\(0.800 \pm 0.000\) &
\(0.603 \pm 0.055\) &
\(0.289 \pm 0.077\) &
\(0.200 \pm 0.000\) &
\(0.213 \pm 0.046\) \\
&
Pattern only &
\(0.379 \pm 0.048\) &
\(0.533 \pm 0.116\) &
\(0.442 \pm 0.070\) &
\(0.289 \pm 0.038\) &
\(0.467 \pm 0.116\) &
\(0.267 \pm 0.023\) \\
&
Identity only &
\(0.500 \pm 0.000\) &
\(0.800 \pm 0.000\) &
\(0.615 \pm 0.000\) &
\(0.267 \pm 0.000\) &
\(0.200 \pm 0.000\) &
\(0.200 \pm 0.000\) \\
&
No Phase~1 &
\(0.489 \pm 0.154\) &
\(0.400 \pm 0.000\) &
\(0.433 \pm 0.058\) &
\(0.156 \pm 0.077\) &
\(0.600 \pm 0.000\) &
\(0.213 \pm 0.046\) \\
\midrule
DWD &
Full &
\(0.567 \pm 0.029\) &
\(0.944 \pm 0.096\) &
\(0.708 \pm 0.042\) &
\(0.181 \pm 0.024\) &
\(0.056 \pm 0.096\) &
\(0.130 \pm 0.016\) \\
&
Pattern only &
\(0.476 \pm 0.041\) &
\(0.611 \pm 0.096\) &
\(0.535 \pm 0.063\) &
\(0.167 \pm 0.000\) &
\(0.389 \pm 0.096\) &
\(0.176 \pm 0.016\) \\
&
Identity only &
\(0.515 \pm 0.079\) &
\(0.833 \pm 0.167\) &
\(0.636 \pm 0.109\) &
\(0.194 \pm 0.024\) &
\(0.167 \pm 0.167\) &
\(0.157 \pm 0.042\) \\
&
No Phase~1 &
\(0.000 \pm 0.000\) &
\(0.000 \pm 0.000\) &
\(0.000 \pm 0.000\) &
\(0.083 \pm 0.000\) &
\(1.000 \pm 0.000\) &
\(0.222 \pm 0.000\) \\
\midrule
Sachs &
Full &
\(0.348 \pm 0.039\) &
\(0.526 \pm 0.053\) &
\(0.417 \pm 0.032\) &
\(0.209 \pm 0.044\) &
\(0.474 \pm 0.053\) &
\(0.231 \pm 0.030\) \\
&
Pattern only &
\(0.378 \pm 0.038\) &
\(0.421 \pm 0.105\) &
\(0.393 \pm 0.054\) &
\(0.147 \pm 0.042\) &
\(0.579 \pm 0.105\) &
\(0.201 \pm 0.021\) \\
&
Identity only &
\(0.366 \pm 0.021\) &
\(0.702 \pm 0.110\) &
\(0.480 \pm 0.044\) &
\(0.253 \pm 0.019\) &
\(0.298 \pm 0.110\) &
\(0.237 \pm 0.005\) \\
&
No Phase~1 &
\(0.361 \pm 0.127\) &
\(0.070 \pm 0.030\) &
\(0.117 \pm 0.049\) &
\(0.026 \pm 0.006\) &
\(0.930 \pm 0.030\) &
\(0.165 \pm 0.008\) \\
\bottomrule
\end{tabular}
\caption{Complete Phase~1 ablation results with GPT-5.4.
All values are mean \(\pm\) sample standard deviation over three runs.}
\label{tab:full-ablation-gpt54}
\end{table*}

\begin{table*}[t]
\centering
\scriptsize
\setlength{\tabcolsep}{3.5pt}
\begin{tabular}{llcccccc}
\toprule
Dataset & Condition &
Precision & Recall & F1 & FPR & FNR & NHD \\
\midrule
Auto-MPG &
Full &
\(0.444 \pm 0.000\) &
\(0.800 \pm 0.000\) &
\(0.571 \pm 0.000\) &
\(0.333 \pm 0.000\) &
\(0.200 \pm 0.000\) &
\(0.240 \pm 0.000\) \\
&
Pattern only &
\(0.444 \pm 0.000\) &
\(0.800 \pm 0.000\) &
\(0.571 \pm 0.000\) &
\(0.333 \pm 0.000\) &
\(0.200 \pm 0.000\) &
\(0.240 \pm 0.000\) \\
&
Identity only &
\(0.415 \pm 0.026\) &
\(0.800 \pm 0.000\) &
\(0.546 \pm 0.022\) &
\(0.378 \pm 0.038\) &
\(0.200 \pm 0.000\) &
\(0.267 \pm 0.023\) \\
&
No Phase~1 &
\(0.476 \pm 0.041\) &
\(0.600 \pm 0.000\) &
\(0.530 \pm 0.026\) &
\(0.222 \pm 0.038\) &
\(0.400 \pm 0.000\) &
\(0.213 \pm 0.023\) \\
\midrule
DWD &
Full &
\(0.524 \pm 0.041\) &
\(0.667 \pm 0.000\) &
\(0.586 \pm 0.025\) &
\(0.153 \pm 0.024\) &
\(0.333 \pm 0.000\) &
\(0.157 \pm 0.016\) \\
&
Pattern only &
\(0.500 \pm 0.000\) &
\(0.667 \pm 0.000\) &
\(0.571 \pm 0.000\) &
\(0.167 \pm 0.000\) &
\(0.333 \pm 0.000\) &
\(0.167 \pm 0.000\) \\
&
Identity only &
\(0.500 \pm 0.000\) &
\(0.667 \pm 0.000\) &
\(0.571 \pm 0.000\) &
\(0.167 \pm 0.000\) &
\(0.333 \pm 0.000\) &
\(0.167 \pm 0.000\) \\
&
No Phase~1 &
\(0.667 \pm 0.000\) &
\(0.333 \pm 0.000\) &
\(0.444 \pm 0.000\) &
\(0.042 \pm 0.000\) &
\(0.667 \pm 0.000\) &
\(0.139 \pm 0.000\) \\
\midrule
Sachs &
Full &
\(0.344 \pm 0.031\) &
\(0.684 \pm 0.105\) &
\(0.456 \pm 0.033\) &
\(0.275 \pm 0.058\) &
\(0.316 \pm 0.105\) &
\(0.256 \pm 0.033\) \\
&
Pattern only &
\(0.250 \pm 0.008\) &
\(0.719 \pm 0.030\) &
\(0.371 \pm 0.013\) &
\(0.451 \pm 0.000\) &
\(0.281 \pm 0.030\) &
\(0.383 \pm 0.005\) \\
&
Identity only &
\(0.332 \pm 0.134\) &
\(0.474 \pm 0.053\) &
\(0.379 \pm 0.109\) &
\(0.245 \pm 0.178\) &
\(0.526 \pm 0.053\) &
\(0.267 \pm 0.134\) \\
&
No Phase~1 &
\(0.338 \pm 0.027\) &
\(0.579 \pm 0.053\) &
\(0.426 \pm 0.030\) &
\(0.238 \pm 0.028\) &
\(0.421 \pm 0.053\) &
\(0.245 \pm 0.019\) \\
\bottomrule
\end{tabular}
\caption{Complete Phase~1 ablation results with GPT-4o.
All values are mean \(\pm\) sample standard deviation over three runs.
For Sachs, cycle reconciliation timed out in all Pattern-only runs and
in one Identity-only run; in those cases, the graph after
bidirectional-edge resolution was evaluated.}
\label{tab:full-ablation-gpt4o}
\end{table*}

\end{appendices}

\end{document}